%% file: main.tex
\documentclass[10pt,twocolumn,letterpaper]{article}

\usepackage[pagenumbers]{cvpr} % To force page numbers, e.g. for an arXiv version

\input{preamble}

\definecolor{cvprblue}{rgb}{0.21,0.49,0.74}
\usepackage[pagebackref,breaklinks,colorlinks,allcolors=cvprblue]{hyperref}

\def\paperID{4759} % *** Enter the Paper ID here
\def\confName{CVPR}
\def\confYear{2025}
\input{math_commands}

\title{\ourname: A Multi-view to Multi-view Diffusion Model \\ for 3D Synthesis and Manipulation}

\author{
Yiftach Edelstein$^1$ \hspace{6mm}  
Or Patashnik$^2$ \hspace{6mm}    Dana Cohen-Bar$^2$ \hspace{6mm} Lihi Zelnik-Manor$^1$ \\[4pt]
$^1$Technion - Israel Institute of Technology	 \hspace{6mm} $^2$Tel Aviv University \hspace{10mm}
\\
 \small\url{https://yiftachede.github.io/Sharp-It/}
         \\[-10pt]
}

\begin{document}
% \maketitle
\input{figures/teaser}
\begin{abstract}
\input{0_abstract}

\end{abstract}

\input{1_intro}

\input{2_related_work}

\input{3_method}
\input{4_experiments}
\input{5_concolusion}
\input{6_acks}

{
    \small
    \bibliographystyle{ieeenat_fullname}
    \bibliography{main}
}


\end{document}

%% file: preamble.tex
\usepackage{multirow}
\usepackage{graphicx}
\usepackage[export]{adjustbox} % for the frame option
\newcommand{\ourname}[0]{Sharp-It}

%% file: math_commands.tex
\usepackage{amsmath,amsfonts,bm}
\usepackage{gensymb}

\def\eqref#1{equation~\ref{#1}}
\def\1{\bm{1}}

\DeclareMathAlphabet{\mathsfit}{\encodingdefault}{\sfdefault}{m}{sl}
\SetMathAlphabet{\mathsfit}{bold}{\encodingdefault}{\sfdefault}{bx}{n}

%% file: figures/teaser.tex
\twocolumn[{%
    \renewcommand\twocolumn[1][]{#1}%
    \maketitle
    \begin{center}
    \vspace{-8pt}
        \includegraphics[width=0.90\textwidth]{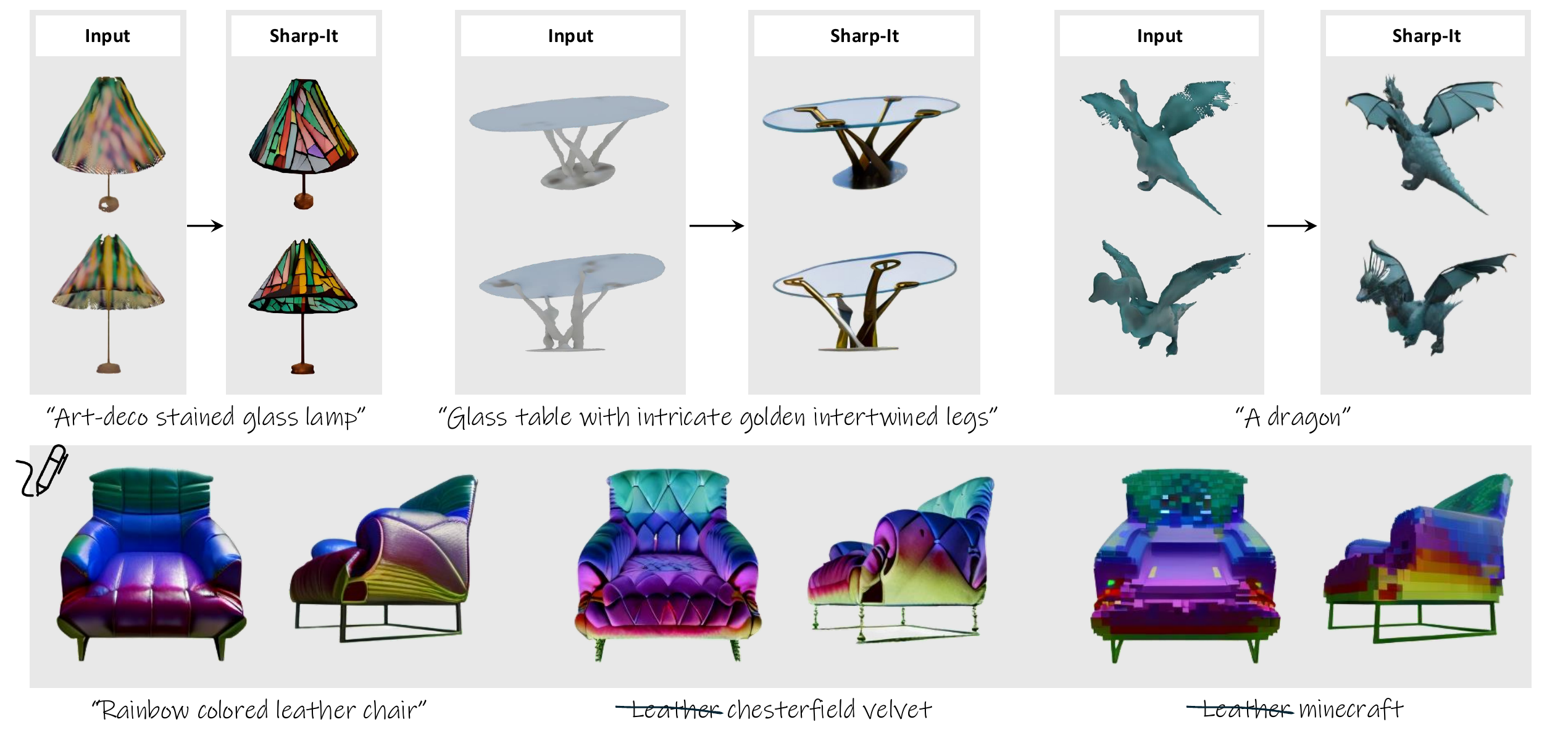}
        % \vspace{-8pt}
        \captionof{figure}{
            \ourname{} is a multi-view to multi-view model that enhances low-quality 3D shapes. It corrects fine-grained geometry details and adds appearance features. 
            The top row displays three degraded shapes and their enhancements by \ourname. 
            The bottom row demonstrates \ourname's capability to edit the appearance of 3D shapes.
        }
        % \vspace{-6pt}
    \label{fig:teaser}
    \end{center}
}] 

%% file: 0_abstract.tex
% opening sentence
Advancements in text-to-image diffusion models have led to significant progress in fast 3D content creation. 
One common approach is to generate a set of multi-view images of an object, and then reconstruct it into a 3D model.
% the problem
However, this approach bypasses the use of a native 3D representation of the object and is hence prone to geometric artifacts and limited in controllability and manipulation capabilities.
An alternative approach involves native 3D generative models that directly produce 3D representations. These models, however, are typically limited in their resolution, resulting in lower quality 3D objects.
% the focus of this work
In this work, we bridge the quality gap between methods that directly generate 3D representations and ones that reconstruct 3D objects from multi-view images.
% how are we doing it
We introduce a multi-view to multi-view diffusion model called \ourname, which takes a 3D consistent set of multi-view images rendered from a low-quality object and enriches its geometric details and texture.
The diffusion model operates on the multi-view set in parallel, in the sense that it shares features across the generated views.
A high-quality 3D model can then be reconstructed from the enriched multi-view set.
% evaluation
By leveraging the advantages of both 2D and 3D approaches, our method offers an efficient and controllable method for high-quality 3D content creation.
We demonstrate that \ourname{} enables various 3D applications, such as fast synthesis, editing, and controlled generation, while attaining high-quality assets.

%% file: 1_intro.tex
\vspace{-8pt}
\section{Introduction}

Creating 3D content plays an important role across industries such as gaming, augmented and virtual reality, and animation. These applications demand efficient, controllable processes for generating high-quality, editable 3D assets. In recent years, text-to-image diffusion models have significantly advanced 3D content creation, introducing new approaches for producing complex, detailed visual assets.

Early approaches to 3D content generation optimized 3D representations by leveraging 2D diffusion models~\cite{poole2022dreamfusion, sjc}, producing high-quality assets but requiring time-consuming per-asset optimization. Recent techniques have adopted a two-stage process: first, synthesizing consistent multi-view images using 2D diffusion models~\cite{liu2023zero1to3, wang2023imagedream, shi2024mvdream, shi2023zero123plus, liu2023one2345, liu2023one2345++}, and second, performing 3D reconstruction from these multi-view images using feed-forward methods~\cite{instant3d2023, xu2024instantmesh}.
This approach accelerates 3D synthesis while maintaining high-quality output. However, by bypassing native 3D representations, it limits controllability and editability of the resulting assets and tends to produce visual artifacts such as the Janus problem and flat objects. Direct 3D generative models~\cite{jun2023shape, nichol2022pointe} offer greater control over creation and editing but are constrained by resolution limitations, resulting in lower-quality assets. This presents a trade-off between the quality of multi-view methods and the controllability of direct 3D generative models.

In this work, we address the quality gap between direct 3D generative models and those that reconstruct 3D objects from multi-view images. To bridge this gap, we introduce \emph{\ourname}, a multi-view to multi-view diffusion model that enhances objects generated by 3D generative models, refining geometric details and adding fine-grained texture features. The high-quality multi-view set produced by \ourname{} can be reconstructed into a 3D model using existing sparse-view feed-forward reconstruction methods~\cite{xu2024instantmesh, jin2024lvsmlargeviewsynthesis, hong2023lrm, zhuang2024gtr}.

\ourname{} is conditioned on a text prompt and operates on the multi-view set in parallel, sharing features across the generated views~\cite{shi2024mvdream, shi2023zero123plus}. We use Shap-E~\cite{jun2023shape} as our backbone generative model, a text-conditioned 3D latent diffusion model that operates on the parameters of implicit functions. Shap-E has been shown to offer rich generative capabilities~\cite{chen2023shapeditor, sella2024spicee}, but produces low-quality assets. To train \ourname, we use pairs of multi-view sets obtained by encoding high-quality 3D objects~\cite{Deitke_2023} into Shap-E's latent space and rendering them.

Our approach offers two key advantages. First, since we start from a coarsely valid 3D object, we inherit its plausible geometric structure, avoiding artifacts like the Janus problem that can occur in direct multi-view synthesis methods. Second, the 3D consistency of the input views facilitates output consistency, allowing the model to focus on generating the fine details necessary for high-quality results.

% The high-quality multi-view set produced by \ourname{} can be reconstructed into a 3D model using existing sparse-view feed-forward reconstruction methods~\cite{xu2024instantmesh, jin2024lvsmlargeviewsynthesis, hong2023lrm, zhuang2024gtr}. Furthermore, we demonstrate that the \ourname-generated multi-view set can serve as keyframes for a video generation model, enabling the creation of 360-degree dense views. 

Our method combines the controllability of native 3D generative models with the capabilities of 2D diffusion models in generating highly detailed images. Through extensive experiments, we demonstrate that \ourname{} outperforms existing enhancement methods in both quality and efficiency. We show various applications, including fast generation of high-quality 3D objects from text prompts and geometric controls~\cite{sella2024spicee}. Additionally, we show that \ourname{} enhances 3D asset editing capabilities.

%% file: 2_related_work.tex
\section{Related Work}

\subsection{Diffusion Models for Content Generation}
Diffusion models~\cite{ho2020denoising, rombach2022highresolutionimagesynthesislatent, saharia2022photorealistictexttoimagediffusionmodels} have demonstrated remarkable success in modeling complex data distributions and generating high-fidelity samples. In the realm of 2D image synthesis, these models have achieved state-of-the-art results, producing photorealistic images with precise control. Motivated by this success, recent works have explored the application of diffusion models to 3D content creation.

\vspace{-14pt}
 \paragraph{3D Diffusion Models}
Some works utilize 3D datasets to train diffusion models for direct 3D asset generation, with different approaches varying in their choice of 3D representations. Some methods employ explicit representations~\cite{zhang2024gaussiancubestructuredexplicitradiance, yariv2024mosaicsdf3dgenerativemodels} such as point clouds~\cite{schröppel2024neuralpointclouddiffusion, nichol2022pointe, zhou20213dshapegenerationcompletion, lyu2023controllablemeshgenerationsparse, NEURIPS2022_40e56dab}. Others transform the input data into learned implicit representations such as triplanes~\cite{shue20223dneuralfieldgeneration,gupta20233dgentriplanelatentdiffusion, chen2023singlestagediffusionnerfunified} or neural networks~\cite{li2023diffusionsdftexttoshapevoxelizeddiffusion, hui2022neuralwaveletdomaindiffusion3d} before learning their distributions. A notable example is Shap-E~\cite{jun2023shape}, which adopts NeRF (Neural Radiance Field) and SDF (Signed Distance Field) as implicit representations and trains a diffusion model to generate the weights of these networks. While these 3D-native diffusion models enable fast generation and provide greater control over the generation process~\cite{sella2024spicee}, they often face challenges due to limited training data and the high computational cost of operating directly on 3D representations.

\vspace{-14pt}
\paragraph{2D Diffusion Models for 3D Generation}
Another approach leverages the powerful prior of pretrained text-to-image diffusion models \cite{rombach2022highresolutionimagesynthesislatent} for 3D generation. Some methods \cite{wang2022scorejacobianchaininglifting, poole2022dreamfusion,metzer2022latent, katzir2024noisefree, chen2023fantasia3d, wang2023prolificdreamer, lin2023magic3dhighresolutiontextto3dcontent, yi2024gaussiandreamerfastgenerationtext,tang2024dreamgaussiangenerativegaussiansplatting} utilize these models to guide the optimization of 3D representations, such as NeRF or Gaussian Splatting, typically through Score Distillation Sampling (SDS)~\cite{poole2022dreamfusion} or its variants.
To enhance and accelerate SDS-guided optimization, recent methods fine-tune 2D diffusion models on 3D datasets~\cite{deitke2023objaversexluniverse10m3d}. Zero-1-to-3~\cite{liu2023zero1to3} fine-tunes a pre-trained diffusion model by incorporating input views and camera parameters to enable novel-view synthesis. Building upon this fine-tuning strategy, some followup works~\cite{shi2024mvdream, liu2024syncdreamergeneratingmultiviewconsistentimages,Szymanowicz_2023_ICCV,shi2023zero123singleimageconsistent} enable simultaneous generation of multiple views using shared attention mechanisms. While these 3D-aware fine-tuning approaches strengthen the prior and improve both convergence speed and result quality when used in SDS-based optimization, the per-shape optimization process still requires substantial computational time.

To overcome the high computational costs of per-shape optimization, recent works propose two-stage approaches that first generate multi-view images and then apply fast mesh reconstruction techniques to create 3D objects from these images. One-2-3-45~\cite{liu2023one2345} and One-2-3-45++~\cite{liu2023one2345++} employ feed-forward networks to perform fast, high-quality asset creation. Wonder3D~\cite{long2023wonder3dsingleimage3d} introduces a cross-domain attention mechanism that generates both normal map images and color images, which facilitates fast mesh reconstruction through novel geometry-aware normal fusion algorithms. 
InstantMesh~\cite{xu2024instantmesh} is a sparse-view 3D reconstruction method that utilizes a feed-forward network based on the LRM~\cite{hong2024lrmlargereconstructionmodel} architecture. Recent advancements in feed-forward sparse-view 3D reconstruction further improve reconstruction quality with different architectures and training paradigms~\cite{zhuang2024gtr, jin2024lvsmlargeviewsynthesis}. While these methods successfully generate high-quality objects rapidly, the reliance on multi-view images limits controllability and editing capabilities.

\subsection{Controlled Generation and Editing}
Extensive work has focused on developing methods for controlled generation and editing of images and video by employing 2D diffusion models~\cite{mokady2023null, garibi2024renoise, Avrahami_2022_CVPR, wu2024turboedit, deutch2024turboedittextbasedimageediting, huberman2024edit, brooks2022instructpix2pix, zhang2023adding, li2023gligen, avrahami2023spatext,cohen2024sliceditzeroshotvideoediting}. Many of these approaches leverage the semantic features learned by the models to gain control and perform edits~\cite{zhang2023adding, hertz2022prompt, Tumanyan_2023_CVPR, patashnik2023localizing, Cao_2023_ICCV, parmar2023zero}.
Some methods have extended the control and editing capabilities of 2D diffusion models to 3D representations~\cite{haque2023instructnerf2nerfediting3dscenes, koo2024posteriordistillationsampling, patashnik2024consolidating, li2023diffusionsdftexttoshapevoxelizeddiffusion, bhat2023loosecontrol, mvedit2024}. However, these typically require computationally expensive optimization processes per object. Faster control and editing of 3D shapes can be achieved through the latent spaces of native 3D generative models~\cite{sella2024spicee, sella2023vox, chen2023shapeditor, hu2023neuralwaveletdomaindiffusion3d, liu2023meshdiffusionscorebasedgenerative3d}. Yet, the quality of shapes generated by these methods is often limited by the output quality of the underlying 3D models, which can be constrained by resolution limitations.
In this work, we propose a novel approach that combines 3D generative models with a multi-view enhancement method to obtain high-quality 3D controlled generation and editing pipelines.

%% file: 3_method.tex
\section{Method}
\input{figures/method}
In this section, we present our approach to addressing the quality gap between direct 3D generative models and those that reconstruct 3D objects from multi-view images.
We focus on Shap-E~\cite{jun2023shape}, a 3D generative model, and introduce \emph{\ourname}, a multi-view-to-multi-view diffusion model that enhances 3D objects generated by Shap-E. We train \ourname{} to improve these objects by adding intricate appearance details and correcting geometric artifacts such as discontinuities and broken parts.
% We begin by providing the necessary background on Shap-E and Zero123++, the diffusion models upon which \ourname{} builds.
The process of generating a 3D model with \ourname{} is demonstrated in Figure~\ref{fig:method}.

\subsection{Preliminaries}

\paragraph{Shap-E} 
is a latent diffusion model specifically designed for generating 3D assets. As common in latent diffusion models~\cite{rombach2022highresolutionimagesynthesislatent}, Shap-E is trained in two stages. In the first stage, an encoder is trained to map 3D objects into a latent space. This latent space corresponds to the weight space of implicit functions that represent 3D shapes, with each shape represented as an element in $\mathbb{R}^{1024\times1024}$. The latent representation can be decoded using Signed Texture Field (STF) rendering, where it is treated as the weights of an implicit function.
In the second stage, a diffusion model is trained within the Shap-E latent space, allowing for conditioning on either text or images.

\vspace{-14pt}
\paragraph{Zero123++}
is an image-conditioned diffusion model designed to generate 3D-consistent multi-view images from a single input view~\cite{shi2023zero123singleimageconsistent}. It builds upon Stable Diffusion~\cite{rombach2022highresolutionimagesynthesislatent}, which is a latent diffusion model comprising a VAE and a UNet. The VAE encodes images into a resolution that is eight times smaller and consists of four channels, while the UNet serves as the diffusion model, operating on the four-channel latent codes.
Zero123++ is a fine-tuned version of Stable Diffusion that accepts an image as input. Given an input image, it produces a $3\times2$ grid of $320\times320$ pixel images, with six constant azimuth and elevation angles. 
Originally, Zero123++~\cite{shi2023zero123singleimageconsistent} was designed to generate this grid image with a grey background. In InstantMesh~\cite{xu2024instantmesh}, it was further fine-tuned to render a white background, addressing the issue of ``floaties''—particles floating through space during 2D-to-3D lifting.

\subsection{Dataset Construction} \label{sec:data}
We begin by constructing a paired dataset consisting of degraded Shap-E objects along with corresponding high-quality objects. Our key idea in constructing the dataset is that such pairs can be obtained by employing the encoder of Shap-E in conjunction with a high-quality 3D objects dataset.
Specifically, we utilize objects from Objaverse~\cite{deitke2023objaversexluniverse10m3d}, provided by~\cite{luo2023scalable, luo2024view}, and encode them using Shap-E's encoder. For each pair of an object from Objaverse and its encoded Shap-E latent code, we render a $3\times2$ grid from six predefined camera views, applying three HDR lighting conditions. Our experiments, detailed in the ablation studies, indicate that rendering each object under varying HDR lighting enhances the model’s performance.

We apply a filtering criteria on the resulting dataset.
We remove objects where the degraded Shap-E rendering is significantly different from the original object's rendering, indicating a failure of Shap-E's encoder. Additionally, we filter out objects that are too thin or do not match certain keywords based on their annotated captions. For each remaining object, we extract a caption using BLIP2~\cite{li2023blip2}.
Finally, we split the dataset into training and test sets, resulting in 180,000 objects for training and 6,000 for testing.

\subsection{\ourname}
\ourname{} is a multi-view to multi-view diffusion model designed to enhance low-quality multi-view images of 3D objects generated by Shap-E. It takes as input a set of multi-view images rendered from a low-quality 3D object, along with a textual prompt, and produces high-quality multi-view images with refined geometric details and textures that correspond to the input views.
% In practice, the multi-view image set is organized as a grid of $3\times2$ views, resulting in a total resolution of $960\times640$. Each view has its own azimuth and elevation angles, which remain fixed across different objects. 
An overview of a 3D generation pipeline with \ourname{} is shown in Figure~\ref{fig:method}.

\paragraph{Architecture}
The architecture of \ourname{} has two key requirements: generating multi-view image sets and incorporating input multi-view sets as conditions. We build on Zero123++~\cite{shi2023zero123plus, xu2024instantmesh}, which was fine-tuned for multi-view generation, where images are arranged in a $3\times2$ grid with a total resolution of $960\times640$ and fixed camera angles across objects.
To enable multi-view conditioning, we modify the architecture of Zero123++ by expanding the UNet input to 8 channels: 4 for latent noise and 4 for the VAE-encoded Shap-E multi-view images. This design lets our model leverage the coarse geometry from the input views to achieve better 3D consistency, and is inspired by image editing techniques that fine-tune diffusion models to accept an image as input and modify specific parts while preserving others~\cite{brooks2022instructpix2pix, rombach2022highresolutionimagesynthesislatent, yang2022paint}. Unlike these approaches, which operate on a single image, our model learns to enhance the input views in a 3D consistent manner. 
Furthermore, we replace Zero123++'s image embedding with text prompts in the cross-attention layers, enabling better enhancement control and appearance editing capabilities (Section~\ref{sec:app-edit}).

The model we build on consists of self-attention layers, which play an important role in facilitating the consistency of our produced multi-view set~\cite{shi2024mvdream, wang2023imagedream}. Since our model operates on a multi-view image grid, these layers can be seen as an application of cross-view attention between the different views. Similarly to previous works~\cite{wang2023imagedream, shi2024mvdream, shi2023zero123plus}, this allows our model to simultaneously refine corresponding points across different views by learning the correspondences between them. 
We visualize the learned correspondences in Figure~\ref{fig:cross-view-attention}. The figure displays self-attention maps for a query point marked by a red dot in the leftmost image. The results demonstrate that this point on the wheel receives the highest attention weight across different views. Additionally, the attention mechanism identifies semantically similar points -- notably, other wheels of the car.

\vspace{-14pt}
\paragraph{Training}

\input{figures/self-attention}

To train \ourname, we utilize our paired dataset consisting of $x$, a highly-detailed multi-view image set; $x_{\text{Shap-E}}$, a degraded version of $x$, representing the low-quality multi-view image generated by Shap-E; and a text prompt $c_{\text{prompt}}$, describing the high-quality 3D object.
We initialize \ourname{} with the weights of Zero123++ taken from the version trained by Xu \etal~\cite{instant3d2023}, and fine-tune it using standard diffusion training with v-prediction. The training loss is formulated as follows:
\[
\mathcal{L} = \mathbb{E}_{t, \epsilon \sim \mathcal{N}(0,1)} \left[ \| v - v_\theta(x_t, x_{\text{Shap-E}} \, ,c_{\text{prompt}}) \|^2 \right],
\]
where $v_\theta$ denotes the v-prediction of the model, parameterized by $\theta$. $x_t$ is obtained by adding noise to $x$ with respect to the diffusion timestep $t$. $t$ and $\epsilon$ are randomly sampled diffusion step and Gaussian noise, respectively. $v$ is defined as $\alpha_t\epsilon - \sigma x$, where $\alpha_t, \sigma$ are parameters of the noise scheduler.
We train our network for 500,000 steps, with a CFG drop probability of 0.1, and a batch size of 3 multi-views, using a single NVIDIA A6000 GPU. 
% The AdamW optimizer is employed, with a learning rate of $10^{-5}$.

\vspace{-10pt}
\paragraph{Inference}
Given a 3D object produced by Shap-E, we first render it from our six predefined azimuths and elevations to form \( x_{\text{Shap-E}} \). To enhance this multi-view image set, we sample a noisy image $x_T \sim \mathcal{N}(0, 1)$ and concatenate it channel-wise to \( x_{\text{Shap-E}} \) to create an 8-channel input to the model. 
The prompt used at inference time can be either the prompt used to generate the object with Shap-E or any other prompt that fits to the object, where the latter option is used to edit the appearance of the object.
We use our trained \ourname{} to iteratively denoise $x_t$, moving from the complete noisy multi-view image set, $x_T$ at $t=T$ to a clean, high-quality multi-view set $\hat{x}$ at $t=0$. The final output multi-view set $\hat{x}$ balances the information from the prompt $c_{\text{prompt}}$ and the degraded multi-view input $x_{\text{Shap-E}}$.

Our enhanced multi-view set can be reconstructed into a 3D object using any existing feed-forward sparse reconstruction method~\cite{xu2024instantmesh, jin2024lvsmlargeviewsynthesis, zhuang2024gtr, zhang2024geolrm}. 
In our experiments, we use InstantMesh~\cite{xu2024instantmesh} for reconstruction and provide its results in the supplementary materials.

%% file: figures/method.tex
\begin{figure*}
    \centering
    \includegraphics[width=\linewidth]{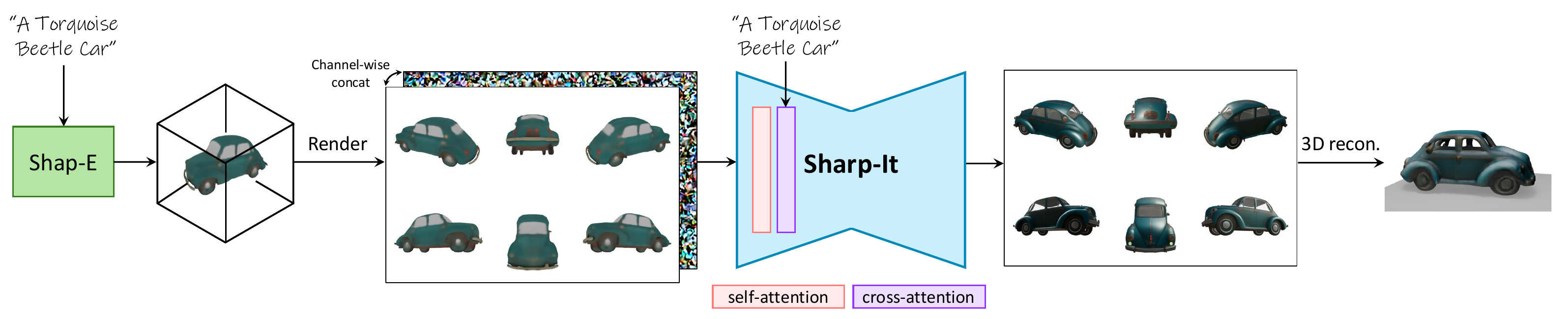}
    \vspace{-18pt}
    \caption{Overview of 3D generation pipeline with \ourname{}.
    First, a 3D object is generated with Shap-E. Then, we render six views of this low-quality object. \ourname{} is a diffusion model based on Stable Diffusion~\cite{rombach2022highresolutionimagesynthesislatent} that enhances these views with the guidance of a text prompt by refining geometry and adding detailed appearance. \ourname{} employs cross-attention layers for text-based guidance and self-attention layers for cross-view consistency. A high-quality 3D object can be reconstructed from the multi-view image set.}
    \vspace{-12pt}
    \label{fig:method}
\end{figure*}

%% file: figures/self-attention.tex
\begin{figure}
    \centering
    \includegraphics[width=0.99\linewidth]{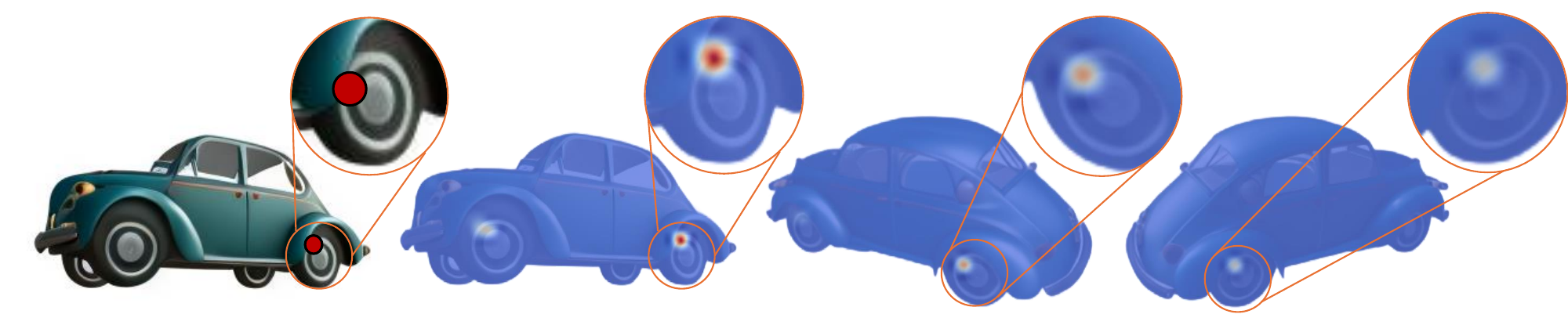}
    \vspace{-10pt}
    \caption{Self-attention maps for a query point (red) on the car's wheel, showing highest attention weights at corresponding wheel locations across different views.}
    \vspace{-14pt}
    \label{fig:cross-view-attention}
\end{figure}

%% file: 4_experiments.tex
\input{figures/comparisons}

\input{tables/experiments}

\section{Experiments}
In this section, we evaluate the performance of \ourname{} in enhancing 3D objects generated by Shap-E. We compare our method against several baselines, conduct ablation studies to analyze the impact of different components, and demonstrate applications in 3D generation and editing.

\subsection{Qualitative and Quantitative Comparison}

\paragraph{Baselines}

We compare \ourname{} with various enhancement baselines. First, we compare against an SDS-based approach that was used in Spice-E~\cite{sella2024spicee}. Specifically, we use GaussianDreamer~\cite{yi2024gaussiandreamerfastgenerationtext}, initializing Gaussian Splatting with Shap-E's shape output and optimizing it with a text-to-image model. 
To make refining the thousands of objects in our test set practical, we limit the optimization time to six minutes per object.
We also include MVEdit~\cite{mvedit2024}, a multi-view editing method that  refines coarse 3D shapes.

\vspace{-10pt}
Another set of baselines combines multi-view image generation with SDEdit~\cite{meng2022sdedit}. Here, we render Shap-E’s shape into a multi-view image set and apply simultaneous edits across views to ensure consistency. Using SDEdit with a strength of 0.4, we add noise to each multi-view image and denoise it. For multi-view generation, we use both MVDream~\cite{shi2024mvdream}, which is guided by a text prompt, and Zero123++~\cite{shi2023zero123plus}, which is conditioned on an image rather than text.
With Zero123++, we test three configurations:
(i) Zero123++ w/ SDEdit (R), where we apply SDXL Refiner model (default strength value)~\cite{podell2023sdxlimprovinglatentdiffusion} on the frontal view of the object, and use this edited view to guide SDEdit across the multi-view set, (ii) Zero123++ w/ SDEdit (C), where SDEdit with Stable Diffusion (strength 0.75) is applied on the object's frontal view which is used to guide the multi-view SDEdit; and (iii) Zero123++ w/ SDEdit (U), where no image condition is used.

\vspace{-10pt}
\paragraph{Dataset}
We conduct our quantitative experiments on our test set described in Section~\ref{sec:data}.
% a subset of Objaverse~\cite{deitke2023objaversexluniverse10m3d} that was held out during the training of \ourname. Each object in this subset is first encoded into Shap-E's latent space and then enhanced using each method. 
To demonstrate our method's generalization capability, we also present results on objects generated directly by Shap-E from text prompts.

\vspace{-12pt}
\paragraph{Metrics}
To assess enhancement performance, we evaluate the image quality produced by each method. Specifically, we compute the FID~\cite{parmar2021cleanfid, heusel2017gans} between the enhanced views and our test set. Additionally, we evaluate the semantic and visual similarity of the enhanced shape with the ground-truth using CLIP~\cite{radford2021learning} and DINO~\cite{caron2021emerging} image encoders.

\vspace{-12pt}
\paragraph{Results}
We present a qualitative comparison in Figure~\ref{fig:mv2mv-comparisons}. The avocado couch and the tiki mask were generated by Shap-E, while the bust is an encoded object from the test set. As shown, our method preserves the objects' coarse details while producing high-quality, detailed enhancements. In contrast, other methods diverge significantly from the original objects and yield less realistic results. For instance, our method accurately generates a leather-like texture, whereas other methods struggle to enhance the flat appearance of the Shap-E-generated couch.
Our approach also maintains color consistency with the input shapes. This is evident in the couch's back section, where brown and green tints are faithfully preserved, as well as in the bust and tiki mask. Our method generates significantly more fine details compared to other methods, particularly visible in the intricate features of the bust and the textural elements of the mask. 
Results for the objects lifted to 3D are provided in the supplementary materials.
% Additional lifted results for each method are provided in the supplementary materials.

Quantitative comparisons are shown in Table~\ref{tab:performance_comparison}. Our method achieves the lowest FID, confirming it produces the highest-quality results. Furthermore, our method shows the best alignment with the ground-truth object, as indicated by the CLIP and DINO similarity metrics, The runtime of our method is comparable to or faster than other methods.

\subsection{Ablation Studies}

\input{figures/ablations_t}

We conduct ablation studies to assess the contributions of different components in \ourname. 
First, we examine the importance of using an input text prompt, which is not utilized in Zero123++~\cite{shi2023zero123plus, xu2024instantmesh}, the model upon which we build.
Second, we explore the effect of constructing the dataset with diverse lighting conditions, rather than using a single type of lighting.
% Specifically, we examine the importance of using an input text prompt and the effect of constructing the dataset with diverse lighting conditions. 

For this analysis, we train our method without each component individually and present the results in Figure~\ref{fig:ablation}.
For computational efficiency, ablation models (including the full method one) were trained for 400,000 steps, fewer than in our main experiments.
\input{tables/ablations}
In the first column, we show a shape generated by Shap-E, which serves as input to the models in the subsequent rows. In the second column, we omit the text prompt, instead using an empty prompt. As shown, without a guiding text prompt, the results lack detail and fail to achieve a metallic appearance. In the third column, we construct the dataset with a single lighting condition, leading to a notable enhancement of the input shape but still lacking finer details. The last column displays results from our full model, which successfully achieves a metallic look and produces highly detailed shapes. Importantly, our results are consistent with the coarse details of the input shape. The lifted object of this example is included in the supplementary materials.
We also quantitatively evaluate each configuration using the metrics described in the previous section. Results are provided in Table~\ref{tab:ablation}. While all configurations are comparable aligned with the ground truth object, the full model achieves the highest image quality.

\input{figures/generation_results}

\subsection{Applications}
We now demonstrate the various applications enabled by \ourname. Specifically, we show how \ourname{} bridges the quality gap between Shap-E and multi-view-based 3D generation methods, while supporting diverse 3D generative applications.

\vspace{-16pt}
\paragraph{Text-to-3D Generation}
As discussed in previous sections, a common approach for text-to-3D synthesis involves first generating a nearly consistent multi-view image set~\cite{shi2024mvdream, wang2023imagedream, shi2023zero123plus}, followed by a sparse-view 3D reconstruction method~\cite{xu2024instantmesh}. By combining Shap-E with \ourname, we achieve text-to-multi-view image synthesis, enabling a complete pipeline for 3D generation (Figure~\ref{fig:method}).
In Figure~\ref{fig:text-to-3d}, we compare text-to-multi-view results from Zero123++ and our method. Our method achieves results comparable to Zero123++ in terms of image quality, while demonstrating superior geometric details. The advantage of using Shap-E as a first stage is particularly evident in the tank and Gundam examples. The tank produced by Zero123++ suffers from a Janus problem, where wheels incorrectly appear on two adjacent sides. Moreover, Zero123++ generates a flat Gundam figure, lacking geometric details. These results demonstrate the effectiveness of our pipeline: using a 3D-aware model to generate a coarse 3D object, followed by refinement using an image diffusion model.

\vspace{-12pt}
\paragraph{3D Object Editing}

\input{figures/edits}
Shap-E provides a latent space in which semantic manipulation of 3D shapes can be performed while maintaining 3D consistency. Our method enhances these edited shapes to achieve high-quality objects. We demonstrate our method's application with two different editing techniques.
First, we use Shap-Editor~\cite{chen2023shapeditor}, which trains a model that takes a Shap-E latent code and a text instruction to produce a new latent code corresponding to the edited shape. While this method enables diverse and fast editing of 3D shapes, it is bounded by Shap-E's quality limitations. \ourname{} addresses this limitation by enhancing the results produced by Shap-Editor.
Second, inspired by diffusion-based editing methods for images, we apply a technique that enables shape editing with Shap-E. Specifically, we apply Edit-Friendly DDPM Inversion~\cite{huberman2024edit} with Shap-E instead of an image diffusion model.

Results for both methods are shown in Figure~\ref{fig:edits}. By combining \ourname{} with existing editing methods, we enable various types of high-quality 3D edits. These include changing an object's shape (demonstrated by transforming a car into an SUV), adding decorations (shown with the Christmas lamp), and modifying colors (as seen with the golden lamp and turquoise car).

\vspace{-16pt}
\paragraph{Appearance Editing} \label{sec:app-edit}
\ourname{} enables detailed control over the appearance of a degraded shape using text prompts. Specifically, at inference time, we can apply a different prompt from the one used during the generation of the Shap-E object, facilitating appearance editing that accurately preserves the original shape. Examples of this application are presented in Figures~\ref{fig:teaser} and \ref{fig:edits-appearance}.

\input{figures/controlled_generation}

\vspace{-14pt}
\paragraph{Controlled Generation}
Previous works have trained models that provide coarse geometric control over objects generated by Shap-E. Similar to Shap-E-based editing methods, such approaches are bounded by Shap-E's quality limitations. We demonstrate high-quality controlled 3D generation by applying our method to shapes generated with Spice-E~\cite{sella2024spicee}.
We show these results in Figure~\ref{fig:controlled-generation}. Our method significantly improves both object textures and fine geometric details. Notably, while Spice-E~\cite{sella2024spicee} uses an SDS-based approach as a refinement step, which impacts its runtime, our method achieves comparable or better results while running an order of magnitude faster.

\input{figures/edits_appearance}

\vspace{-3pt}

%% file: figures/comparisons.tex
\begin{figure*}
    \centering
    \setlength{\tabcolsep}{3pt}
    {\scriptsize
    \begin{tabular}{ccccccccc}

        & Input (Shap-E) & GaussianDreamer & MVEdit & MVDream w/  & Zero123++ w/ & Zero123++ w/ & Zero123++ w/ & Ours (\ourname) \\
        & & & & SDEdit & SDEdit (U) & SDEdit (C)  & SDEdit (R)\\
        %\
        
        \raisebox{43pt}{\multirow{2}{*}{\rotatebox[origin=t]{90}{``An avocado shaped leather couch''}}} &
        \includegraphics[width=0.10\linewidth]{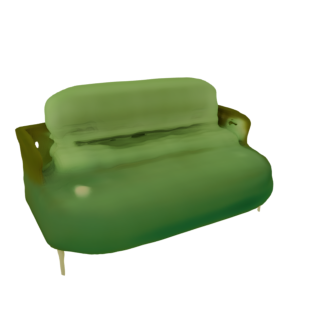} &
        \includegraphics[width=0.10\linewidth, trim=30 30 30 30, clip,]{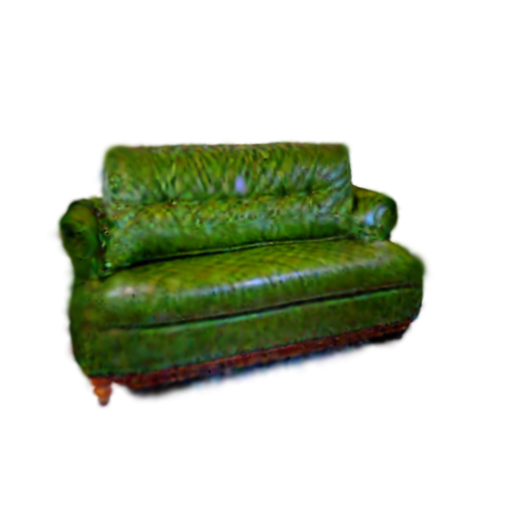} &
        \includegraphics[width=0.10\linewidth]{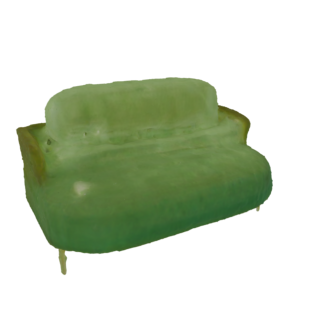} &
        \includegraphics[width=0.10\linewidth]{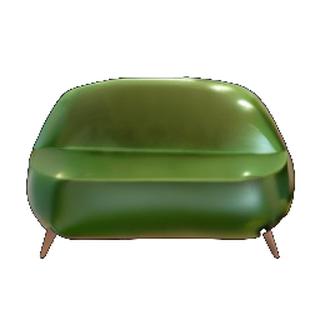} &
        \includegraphics[width=0.10\linewidth]{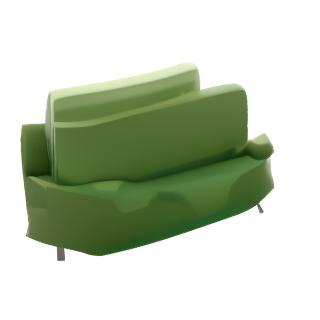} &
        \includegraphics[width=0.10\linewidth]{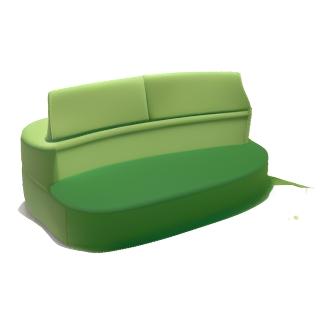} &
         \includegraphics[width=0.10\linewidth]{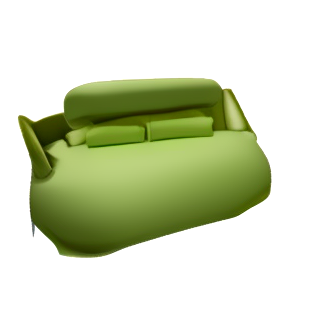} &
        \includegraphics[width=0.10\linewidth]{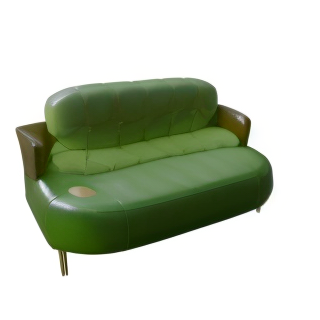} \\
        &
        \includegraphics[width=0.10\linewidth]{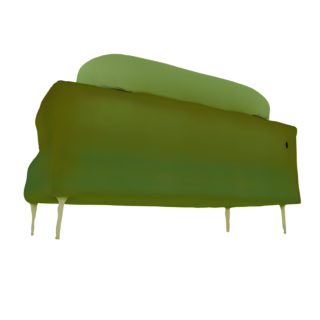} &
       \includegraphics[width=0.10\linewidth, trim=30 30 30 30, clip,]{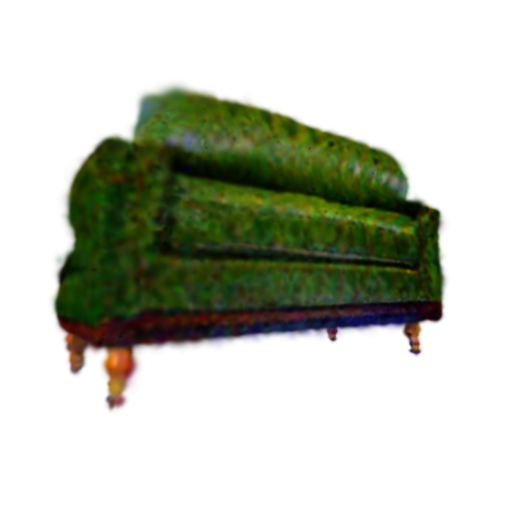} &
        \includegraphics[width=0.10\linewidth]{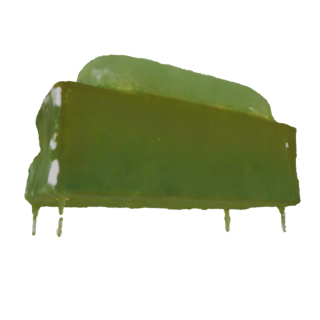} &
        \includegraphics[width=0.10\linewidth]{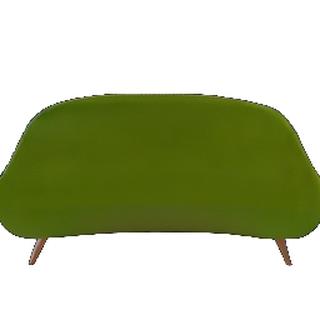} &
        \includegraphics[width=0.10\linewidth]{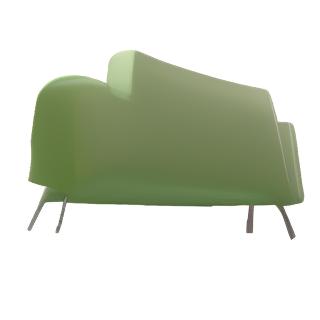} &
        \includegraphics[width=0.10\linewidth]{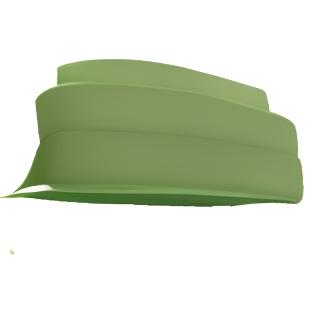} &
        \includegraphics[width=0.10\linewidth]{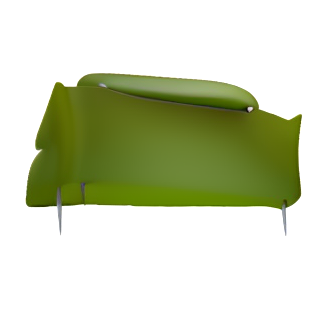} &
        \includegraphics[width=0.10\linewidth]{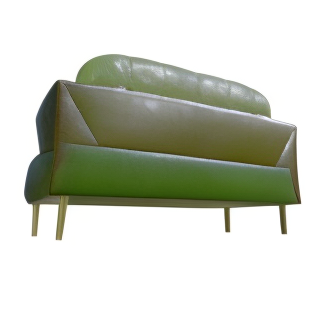} \\
        \raisebox{36pt}{\multirow{2}{*}{\rotatebox[origin=t]{90}{``A bust of a man with a beard''}}} &
        \includegraphics[width=0.10\linewidth]{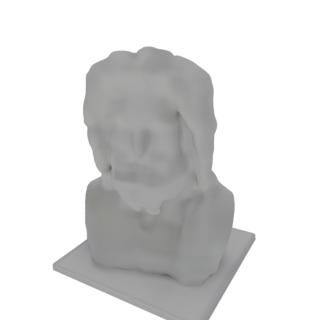} &
        \includegraphics[width=0.10\linewidth]{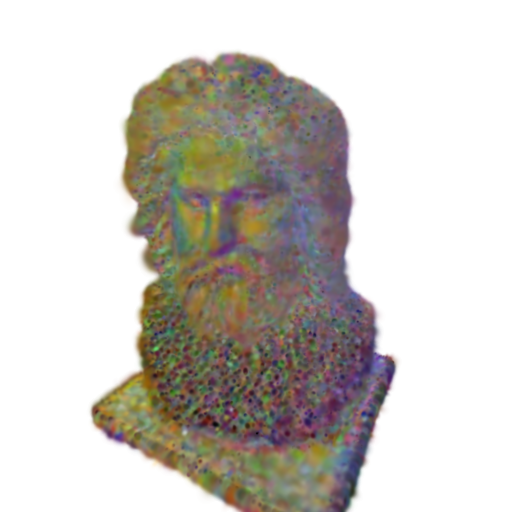} &
        \includegraphics[width=0.10\linewidth]{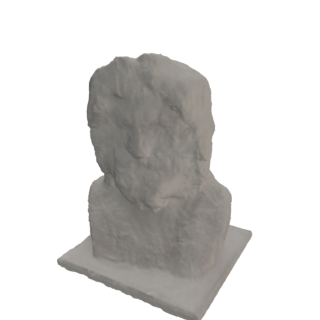} &
        \includegraphics[width=0.10\linewidth]{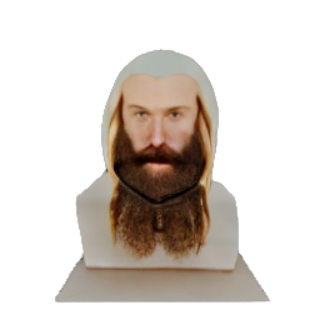} &
        \includegraphics[width=0.10\linewidth]{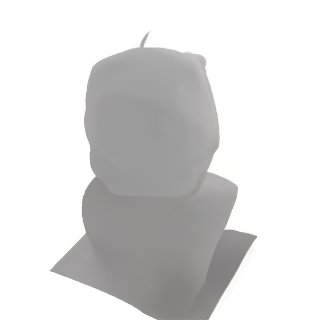} &
        \includegraphics[width=0.10\linewidth]{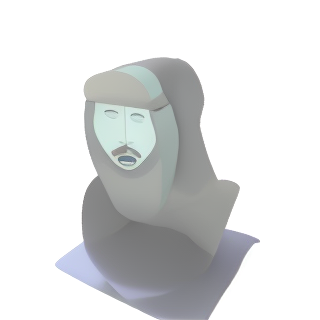} &
        \includegraphics[width=0.10\linewidth]{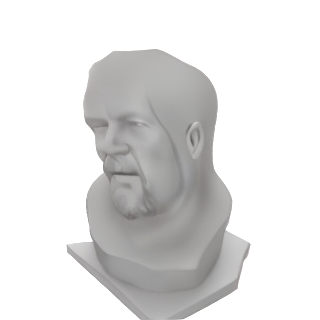} &
        \includegraphics[width=0.10\linewidth]{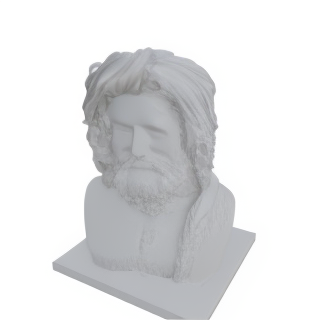} \\
         &
        \includegraphics[width=0.10\linewidth]{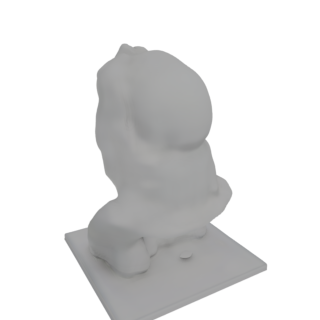} &
        \includegraphics[width=0.10\linewidth]{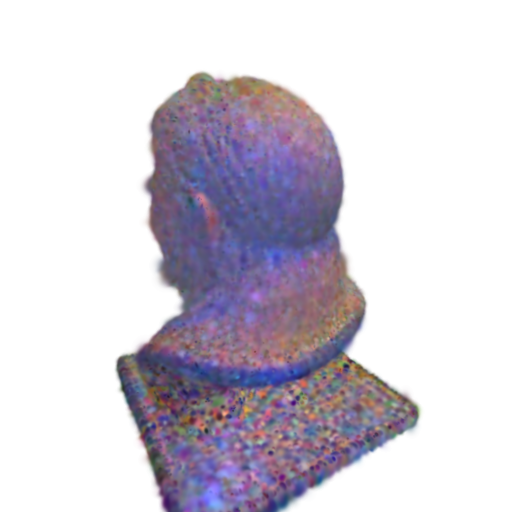} &
        \includegraphics[width=0.10\linewidth]{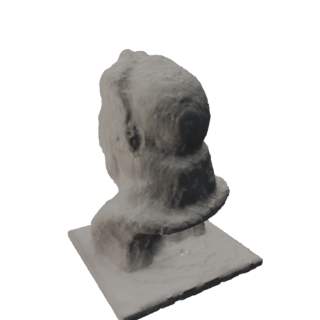} &
        \includegraphics[width=0.10\linewidth]{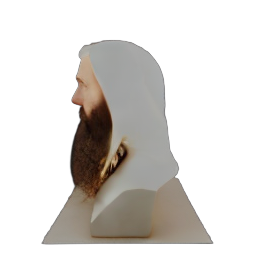} &
        \includegraphics[width=0.10\linewidth]{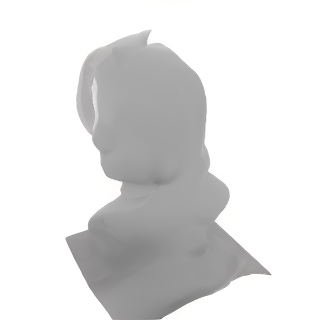} &
        \includegraphics[width=0.10\linewidth]{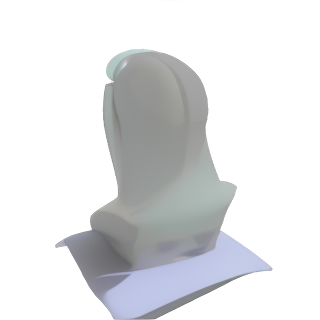} &
         \includegraphics[width=0.10\linewidth]{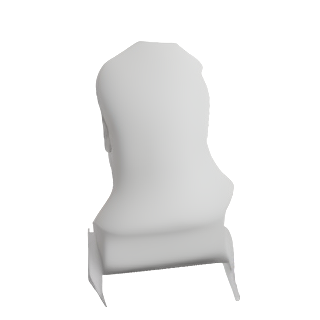} &
        \includegraphics[width=0.10\linewidth]{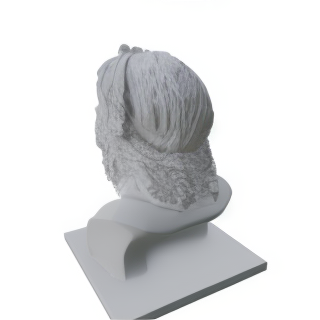} \\
        \raisebox{25pt}{\multirow{2}{*}{\rotatebox[origin=t]{90}{``A wooden tiki mask''}}} &
        \includegraphics[width=0.10\linewidth]{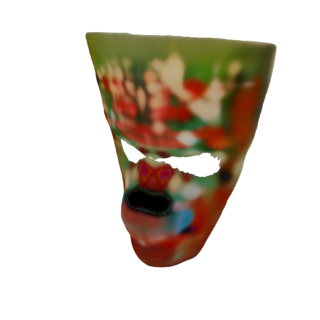} & 
        \includegraphics[width=0.10\linewidth]{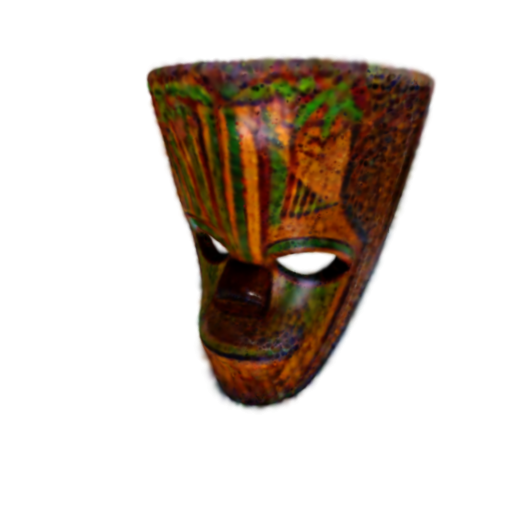} &
        \includegraphics[width=0.10\linewidth]{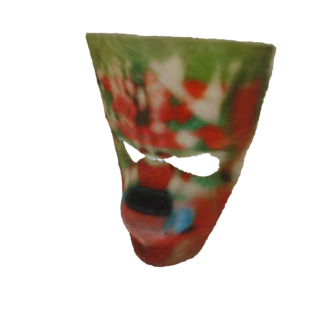} &
        \includegraphics[width=0.10\linewidth]{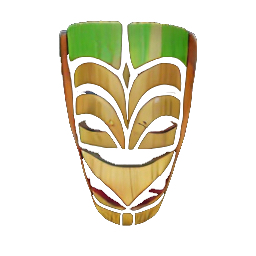} &
        \includegraphics[width=0.10\linewidth]{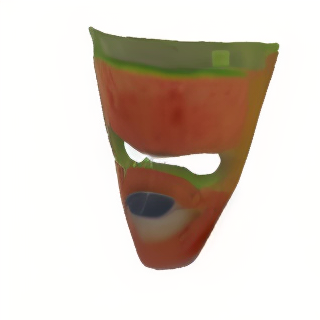} &
        \includegraphics[width=0.10\linewidth]{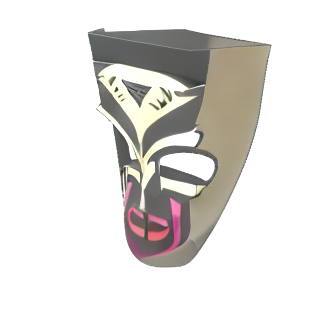} &
         \includegraphics[width=0.10\linewidth]{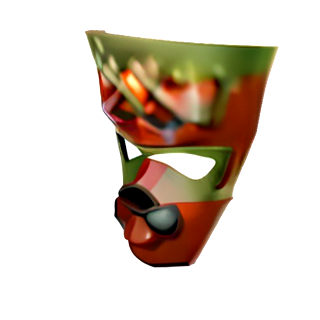} &
        \includegraphics[width=0.10\linewidth]{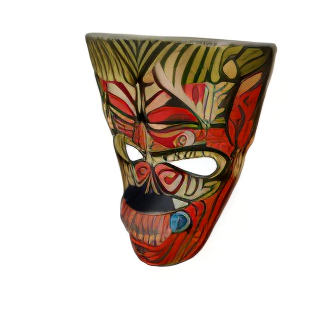} \\
        &
        \includegraphics[width=0.10\linewidth]{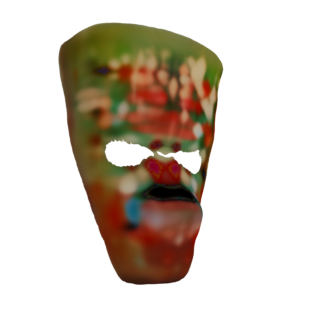} & 
        \includegraphics[width=0.10\linewidth]{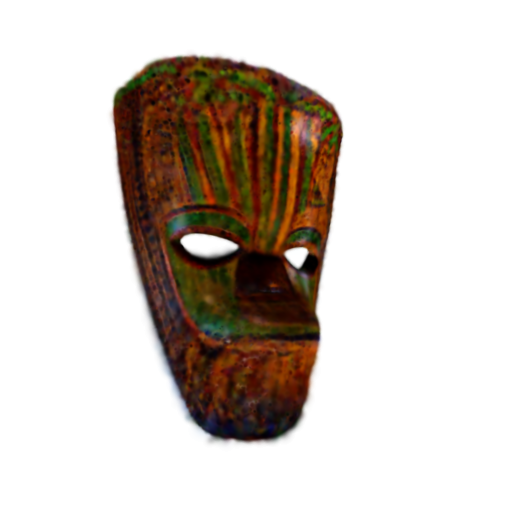} &
        \includegraphics[width=0.10\linewidth]{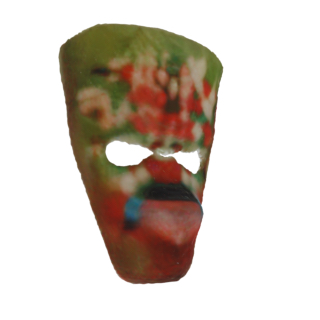} &
        \includegraphics[width=0.10\linewidth]{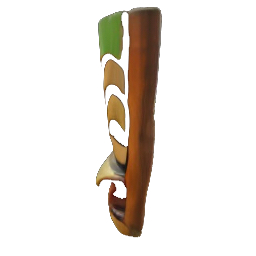} &
        \includegraphics[width=0.10\linewidth]{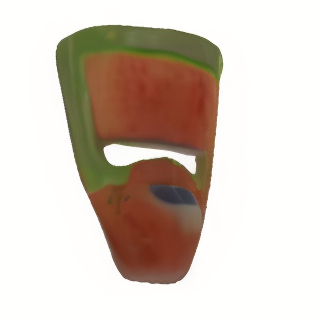} &
        \includegraphics[width=0.10\linewidth]{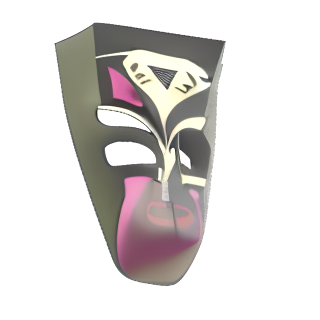} &
        \includegraphics[width=0.10\linewidth]{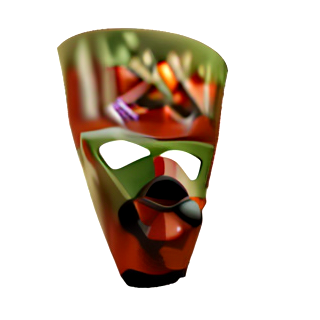} &
        \includegraphics[width=0.10\linewidth]{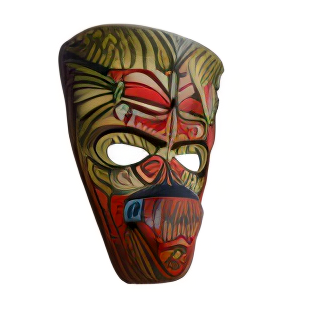} \\

    \end{tabular}
    }
    \vspace{-12pt}
    \caption{
    Comparison of \ourname{} with other methods for 3D object enhancement (GaussianDreamer and MVEdit), and multi-view enhancement (SDEdit based). The first column shows the input object generated by Shap-E. As can be seen, our method achieves the highest quality results while best preserving the input object. 
    }
    \vspace{-12pt}
    \label{fig:mv2mv-comparisons}
\end{figure*}

%% file: tables/experiments.tex
\begin{table}[t]
    \caption{Quantitative comparison of our enhancement method with other baseline methods.}
    \vspace{-10pt}
    \label{tab:performance_comparison}
    \centering
    \setlength{\tabcolsep}{3pt} % Adjusts column padding for narrower width
    {\small % Keeps the font size at small
    \begin{tabular}{@{}lcccc@{}}
        \toprule
        \textbf{Method} & \textbf{FID $\downarrow$} & \textbf{CLIP $\uparrow$} & \textbf{DINO $\uparrow$} & \textbf{Runtime} \\ 
        \midrule
        GaussianDreamer & 50.89 & 0.81 & 0.82 &  6 min \\
        MVEdit & 44.87 & 0.83 & 0.77 &  1 min \\
        MVDream w/ SDEdit & 28.71 & 0.81 & 0.83 &  10 sec \\
        Zero123++ w/ SDEdit (U) & 24.59 & 0.87 & 0.86 &  10 sec \\
        Zero123++ w/ SDEdit (C) & 19.95 & 0.85 & 0.87 &  10 sec \\
        Zero123++ w/ SDEdit (R) & 19.13 & 0.87 & 0.89 &  10 sec \\
        Sharp-It & \textbf{6.60} & \textbf{0.90} & \textbf{0.92} &  10 sec \\
        \bottomrule
    \end{tabular}
    }
    \vspace{-12pt}
\end{table}

%% file: figures/ablations_t.tex
\begin{figure}
    \centering
    \setlength{\tabcolsep}{1pt}
    {\small
    \begin{tabular}{ccccc}
        & Input & W/o text  & W/o diverse & Full method \\
        && prompt & lighting \\
        \raisebox{35pt}{\rotatebox[origin=c]{90}{View 1}} &
        \multicolumn{1}{c}{\includegraphics[width=0.24\linewidth, trim=50 0 50 0, clip]{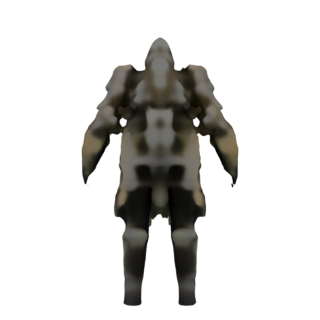}} &
        \includegraphics[width=0.24\linewidth, trim=50 0 50 0, clip]{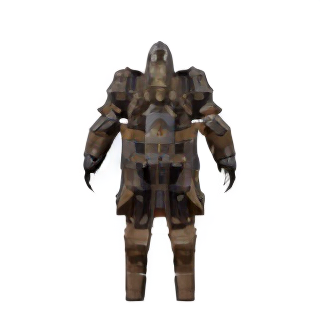} &
        \includegraphics[width=0.24\linewidth, trim=50 0 50 0, clip]{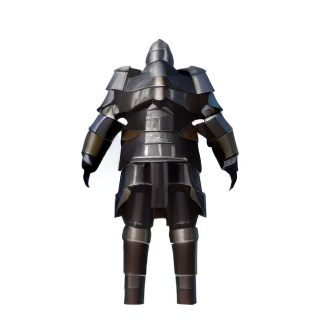} &
        \includegraphics[width=0.24\linewidth, trim=50 0 50 0, clip]{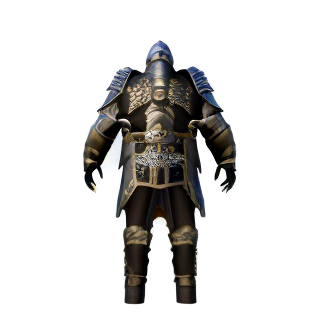} \\
        \raisebox{35pt}{\rotatebox[origin=c]{90}{View 2}} &
        \multicolumn{1}{c}{\includegraphics[width=0.24\linewidth, trim=50 0 50 0, clip]{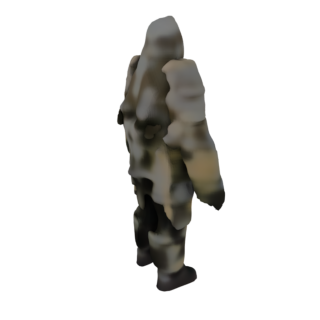}} &
        \includegraphics[width=0.24\linewidth, trim=50 0 50 0, clip]{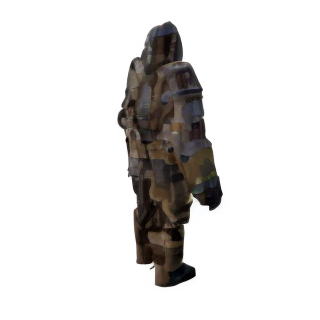} &
        \includegraphics[width=0.24\linewidth, trim=50 0 50 0, clip]{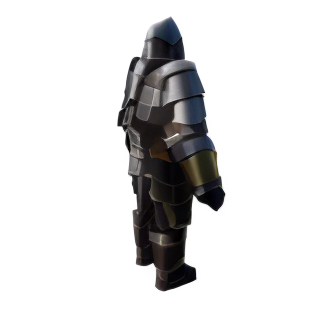} &
        \includegraphics[width=0.24\linewidth, trim=50 0 50 0, clip]{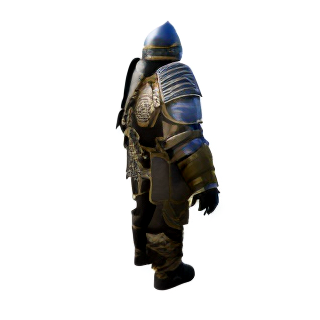} \\
        \raisebox{35pt}{\rotatebox[origin=c]{90}{View 3}} &
        \multicolumn{1}{c}{\includegraphics[width=0.24\linewidth, trim=50 0 50 0, clip]{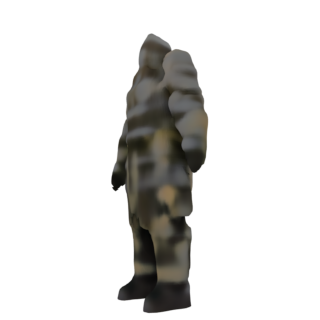}} &
        \includegraphics[width=0.24\linewidth, trim=50 0 50 0, clip]{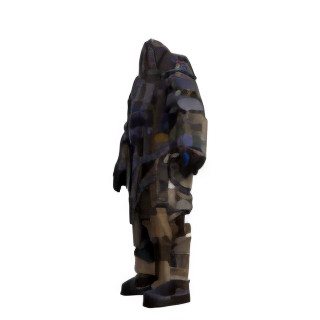} &
        \includegraphics[width=0.24\linewidth, trim=50 0 50 0, clip]{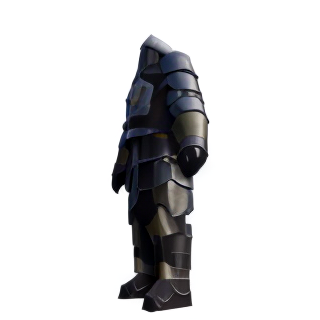} &
        \includegraphics[width=0.24\linewidth, trim=50 0 50 0, clip]{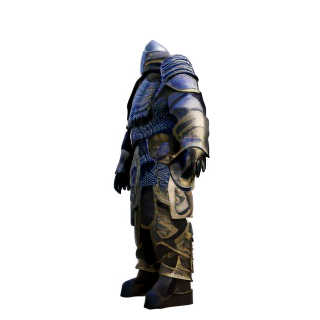} \\
        \multicolumn{5}{c}{``A knight in full plate armor''}
    \end{tabular}
    }
    \caption{Qualitative ablation study. The first column shows the degraded input object generated by Shap-E. Subsequent columns show the effects of removing specific components: omitting the text prompt leads to reduced texture detail, while excluding diverse lighting results in a flatter appearance with less realistic shading. The full model achieves the most refined and detailed result.}
    \label{fig:ablation}
\end{figure}

%% file: tables/ablations.tex
\begin{table}[ht]
    \centering
    \caption{Quantitative results of our ablation study. All models were trained for 400,000 steps.}
    \label{tab:ablation}
    \begin{tabular}{@{}lcccc@{}} % Align columns to the left
        \toprule
        \textbf{Ablation} & \textbf{FID $\downarrow$} & \textbf{CLIP $\uparrow$} & \textbf{DINO $\uparrow$}  \\ 
        \midrule
            W/o text prompt & 9.91 & \textbf{0.91} &0.91&  \\
        W/o diverse lighting  & 9.16 & 0.89 & 0.91 &  \\
        Full method & \textbf{8.06} & 0.89 & 0.91 \\

        \bottomrule
    \end{tabular}
    \vspace{-16pt}
\end{table}

%% file: figures/generation_results.tex
\begin{figure*}
\centering

\setlength{\tabcolsep}{1pt}
{\small
\begin{tabular}{c c c c c c c c c}
    & \multicolumn{2}{c}{``A Gundam''} & { } & \multicolumn{2}{c}{``A tank''} & { } & \multicolumn{2}{c}{``A glass and metal battle-axe''} \\
    \raisebox{20pt}{\rotatebox[origin=c]{90}{Input}} &
    \includegraphics[width=0.13\linewidth]{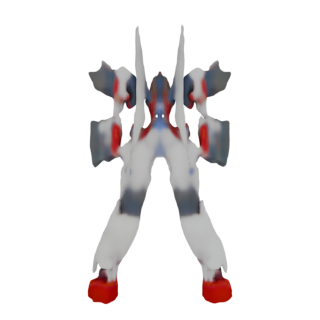} &
    \includegraphics[width=0.13\linewidth]{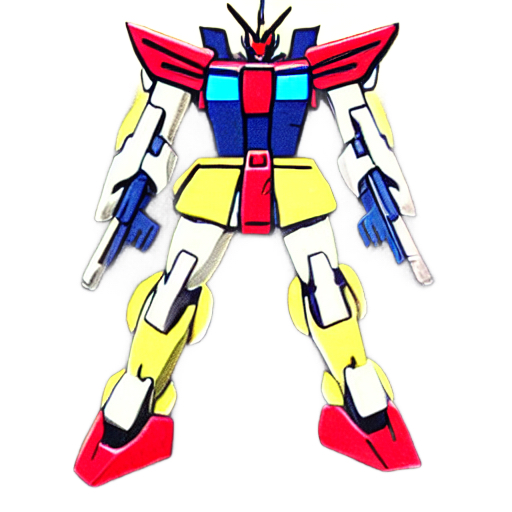} & { } & 
    \includegraphics[width=0.13\linewidth, trim=0 0 0 60, clip]{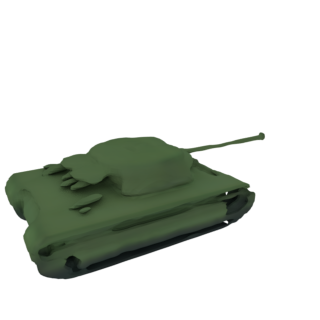} &
    \includegraphics[width=0.13\linewidth, trim=0 50 0 150, clip]{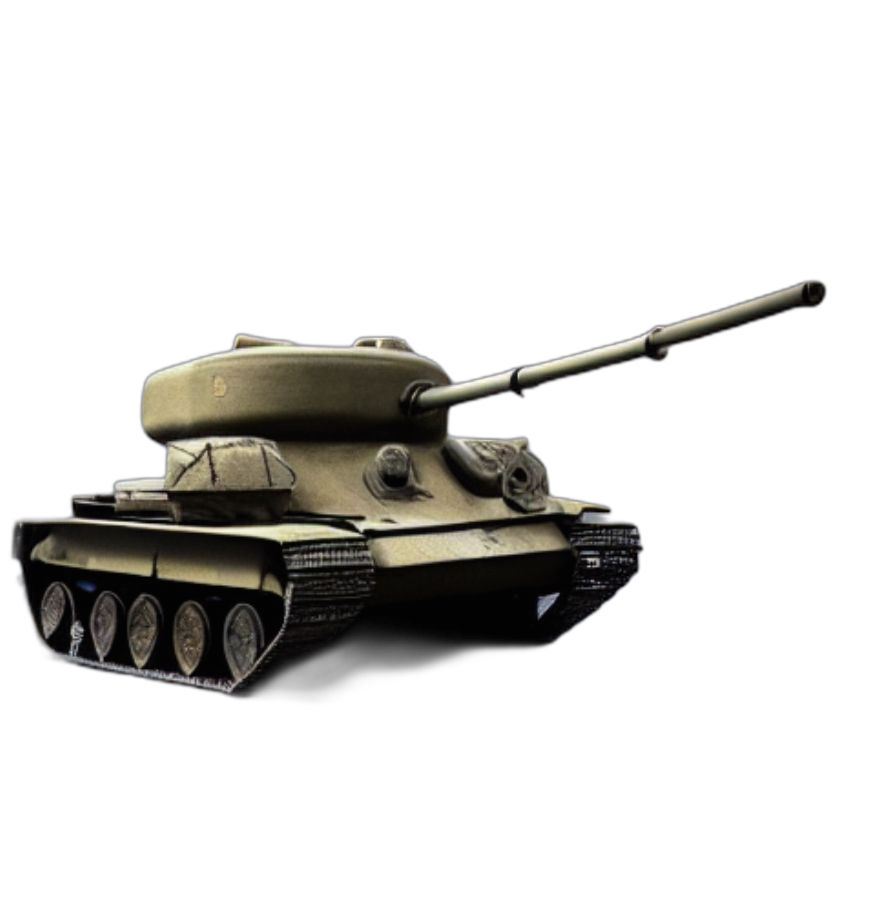} & { } & 
    \includegraphics[width=0.13\linewidth]{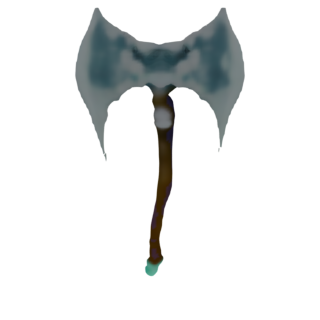} &
    \includegraphics[width=0.13\linewidth]{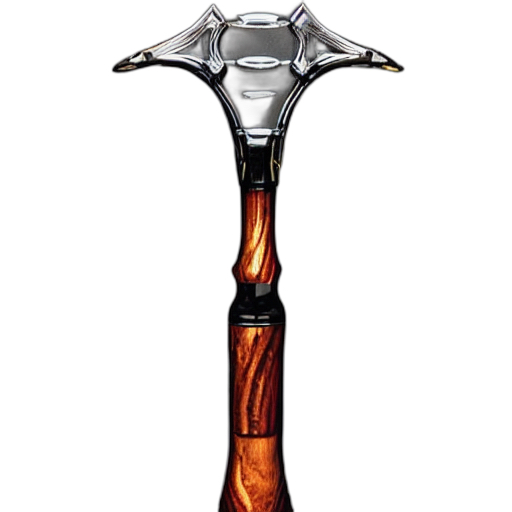}
    \\
      \raisebox{20pt}{\rotatebox[origin=c]{90}{View 1}} &
    \includegraphics[width=0.13\linewidth]{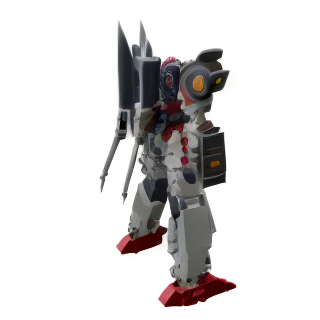} &
    \includegraphics[width=0.13\linewidth]{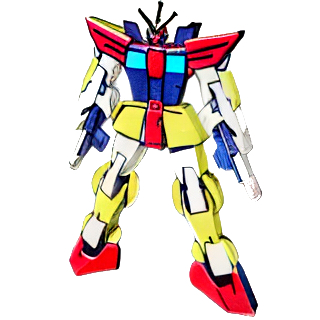} & { } & 
    \includegraphics[width=0.13\linewidth, trim=20 50 20 50, clip]{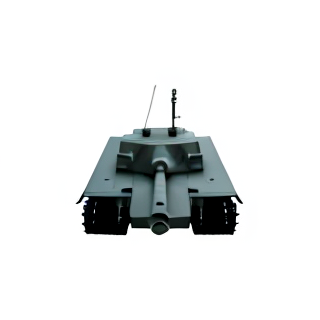} &
    \includegraphics[width=0.13\linewidth, trim=0 20 0 20, clip]{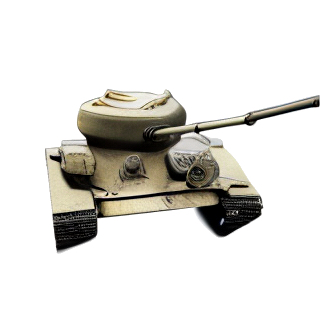} & { } & 
    \includegraphics[width=0.13\linewidth]{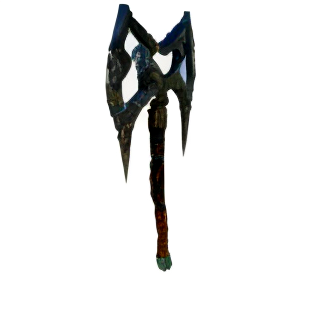} &
    \includegraphics[width=0.13\linewidth]{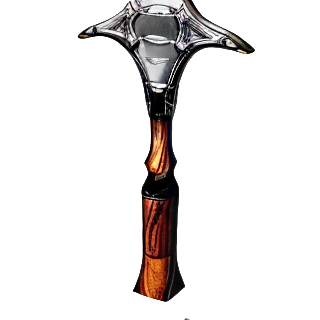}
    \\
    \raisebox{20pt}{\rotatebox[origin=c]{90}{View 2}} &
    \includegraphics[width=0.13\linewidth]{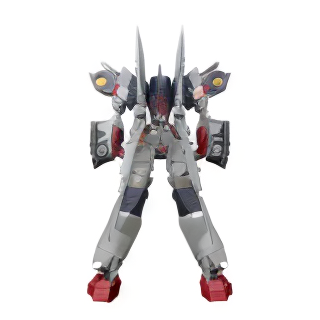} &
    \includegraphics[width=0.13\linewidth]{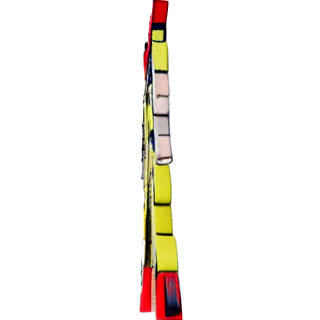} & { } & 
    \includegraphics[width=0.13\linewidth, trim=0 0 0 60, clip]{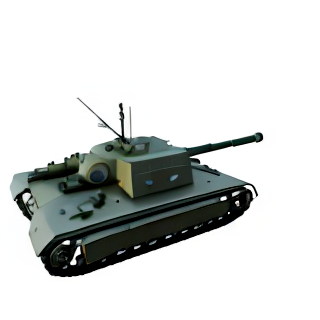} &
    \includegraphics[width=0.13\linewidth, trim=0 0 0 20, clip]{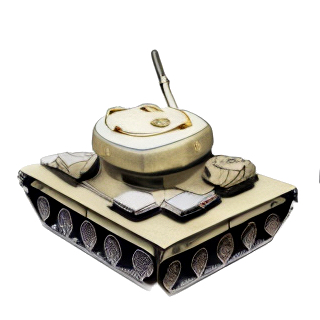} & { } & 
    \includegraphics[width=0.13\linewidth]{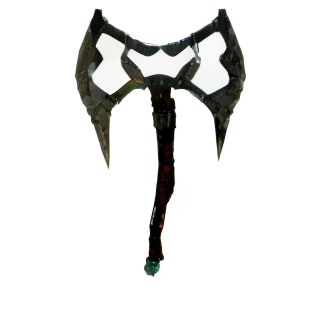} &
    \includegraphics[width=0.13\linewidth]{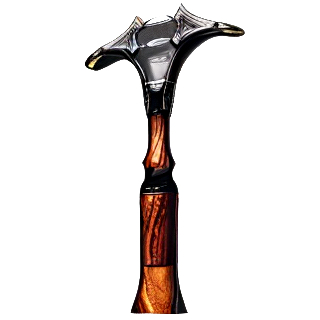} \\
    & \multicolumn{1}{c}{Ours} & \multicolumn{1}{c}{Zero123++} & { } & \multicolumn{1}{c}{Ours} & \multicolumn{1}{c}{Zero123++} & { } & \multicolumn{1}{c}{Ours} & \multicolumn{1}{c}{Zero123++} \\
     
\end{tabular}
}
\vspace{-8pt}
\caption{
Text-to-MultiView qualitative results. Our method generates a multi-view set by creating a 3D object with Shap-E and refining it with \ourname. In contrast, Zero123++ first generates a single image with Stable Diffusion, then produces a multi-view set conditioned on that image. This approach leads to geometric artifacts, such as flatness (gundam) and the Janus problem (tank). Leveraging Shap-E's 3D knowledge, our method yields high-quality, coherent objects. Additional 3D renderings of these results are provided in the supplemental.
}
\vspace{-12pt}
\label{fig:text-to-3d}
\end{figure*}

%% file: figures/edits.tex
\begin{figure}
    \centering
    \setlength{\tabcolsep}{1pt}
    {\scriptsize
    \begin{tabular}{ccccc}
        & \multicolumn{4}{c}{``A table lamp'' $\longrightarrow$ ``A golden table lamp''} \\
        \raisebox{12pt}{\multirow{3}{*}{\rotatebox[origin=t]{90}{Shap-Editor}}} &
                \includegraphics[width=0.15\linewidth, trim=70 45 70 40, clip]
                % L B R T
{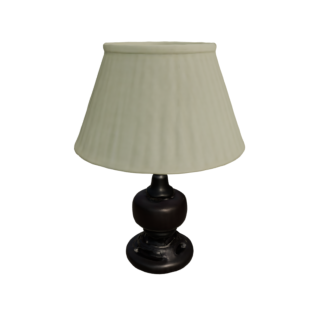} &
                       \includegraphics[width=0.15\linewidth, trim=70 45 70 40, clip]
{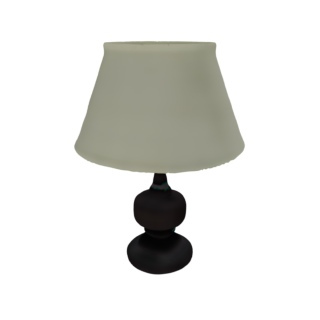} &
                        \includegraphics[width=0.15\linewidth, trim=70 45 70 40, clip]
{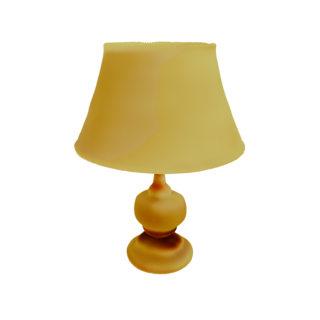} &
                       \includegraphics[width=0.15\linewidth, trim=70 45 70 40, clip]
{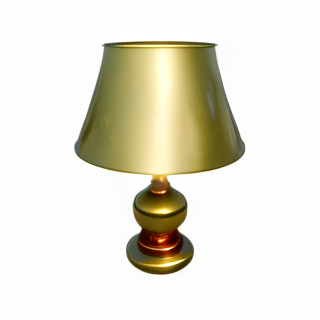} \\
        & \multicolumn{4}{c}{``A table lamp'' $\longrightarrow$ ``A santa table lamp''} \\
         &
                       \includegraphics[width=0.15\linewidth, trim=70 45 70 40, clip]
{images/editings/shap-editor/lamps/original/render_0000.png} &
                      \includegraphics[width=0.15\linewidth, trim=70 45 70 40, clip]
{images/editings/shap-editor/lamps/shap_e/original_shap_e_tile_0.png} &
                    \includegraphics[width=0.15\linewidth, trim=70 45 70 40, clip]
{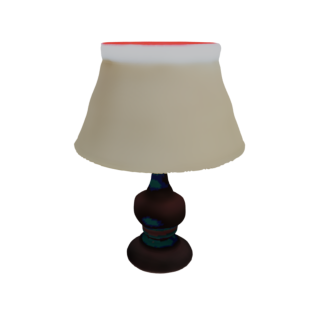} &
                      \includegraphics[width=0.15\linewidth, trim=70 45 70 40, clip]
{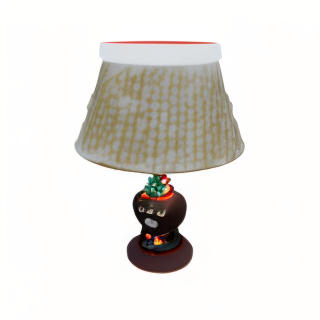} \\
        & \multicolumn{4}{c}{``A blue beetle car'' $\longrightarrow$ ``A torquoise beetle car''} \\
        %
                        % L B R T

        \raisebox{12pt}{\multirow{3}{*}{\rotatebox[origin=t]{90}{DDPM Inversion}}} &
    \includegraphics[width=0.22\linewidth, trim=30 60 15 85, clip]
{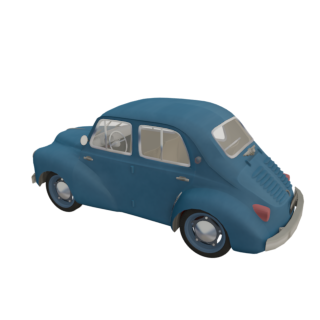} &
           \includegraphics[width=0.22\linewidth, trim=30 60 15 85, clip]{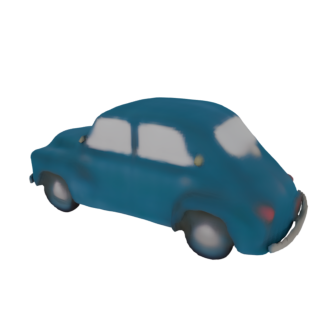} &
           \includegraphics[width=0.22\linewidth, trim=30 60 0 85, clip]{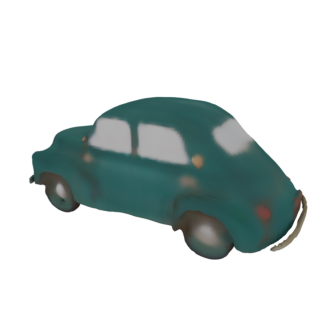} &
           \includegraphics[width=0.22\linewidth, trim=30 60 0 85, clip]{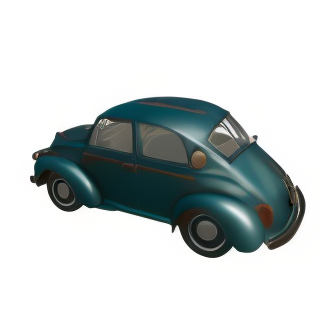} \\
         & \multicolumn{4}{c}{``A blue beetle car'' $\longrightarrow$ ``A blue SUV''} \\
         &
             \includegraphics[width=0.22\linewidth, trim=30 60 15 85, clip]{images/editings/ddpm_inv/original/render_0000.png} &
            \includegraphics[width=0.22\linewidth, trim=30 60 15 85, clip]{images/editings/ddpm_inv/shap_e/original_shap_e_tile_0.png} &
           \includegraphics[width=0.22\linewidth, trim=30 60 0 85, clip]{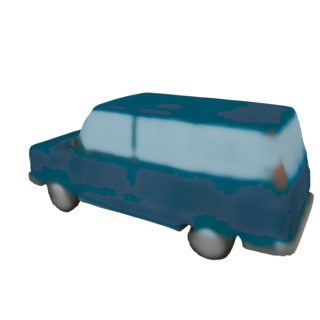} &
               \includegraphics[width=0.22\linewidth, trim=30 60 0 85, clip]{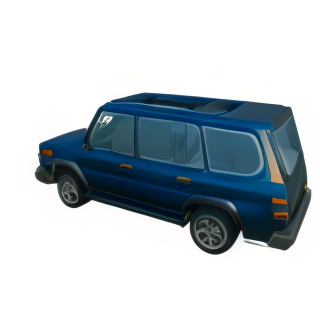} \\
        & Input Shape & Encoded Shaped & Edit & + \ourname{}
    \end{tabular}
    }
    \vspace{-4pt}
    \caption{
    Our method enhances edits performed in Shap-E's space (rightmost column). We show editing results obtained with an existing editing method (Shap-Editor), and demonstrate that DDPM Inversion, originally developed for image diffusion models, works with Shap-E and can be integrated with \ourname{}.
    }
    \vspace{-16pt}
    \label{fig:edits}
\end{figure}

%% file: figures/controlled_generation.tex
\begin{figure}
    \centering
    \setlength{\tabcolsep}{1pt}
    {\small
    \begin{tabular}{cccc}
        \multicolumn{4}{c}{``A red velvet chesterfield chair''} \\
                \includegraphics[width=0.24\linewidth, trim=20 20 40 40, clip]{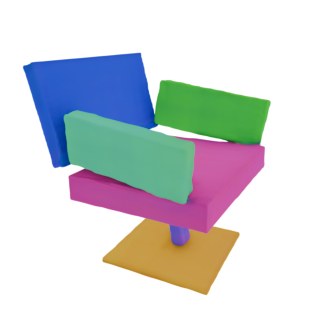} &
                \includegraphics[width=0.24\linewidth, trim=40 40 60 60, clip]{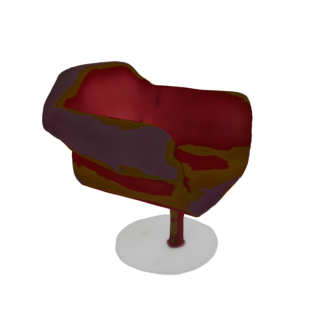} &

                \includegraphics[width=0.24\linewidth, trim=60 60 100 90, clip]{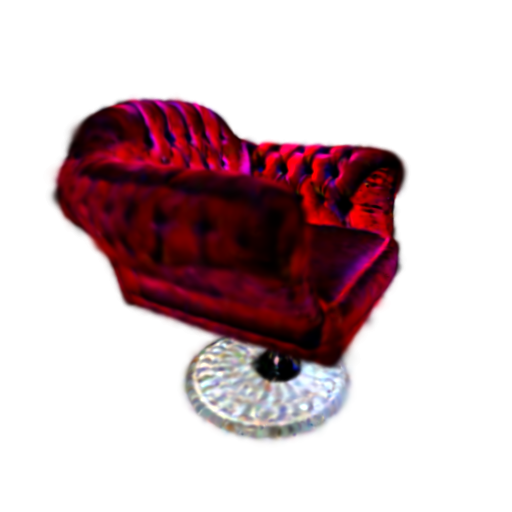} &
                \includegraphics[width=0.24\linewidth, trim=40 40 60 60, clip]{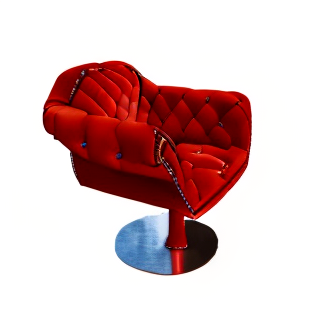} \\
                \includegraphics[width=0.24\linewidth, trim=40 50 40 60, clip]{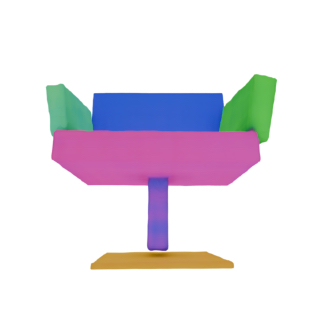} &
                \includegraphics[width=0.24\linewidth, trim=40 50 40 60, clip]{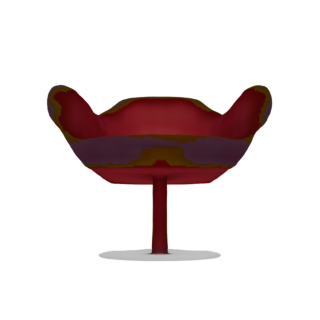} &
                \includegraphics[width=0.24\linewidth, trim=50 60 50 90, clip]{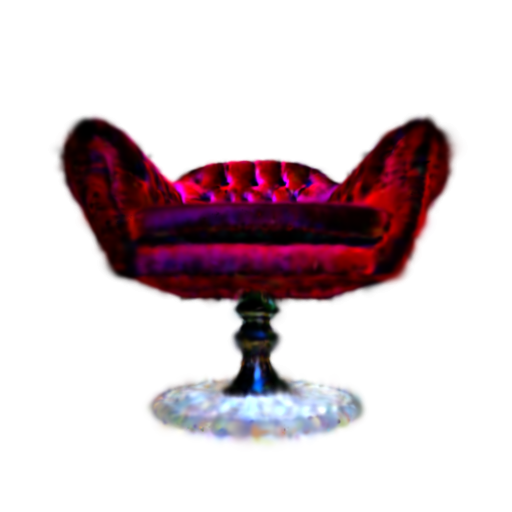} &
                \includegraphics[width=0.24\linewidth, trim=40 50 40 60, clip]{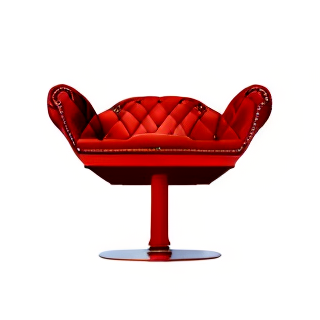} \\

        \multicolumn{4}{c}{``A cargo spaceship''} \\
                \includegraphics[width=0.24\linewidth, trim=60 60 20 50, clip]{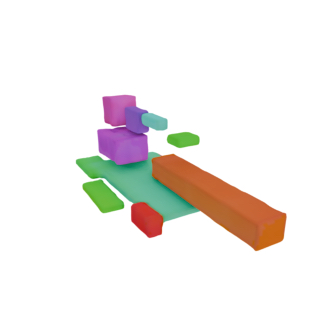} &
                \includegraphics[width=0.24\linewidth, trim=60 60 20 50, clip]{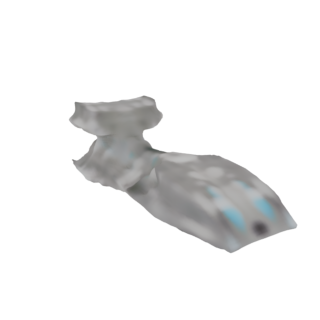} &
                \includegraphics[width=0.24\linewidth, trim=60 60 20 105, clip]{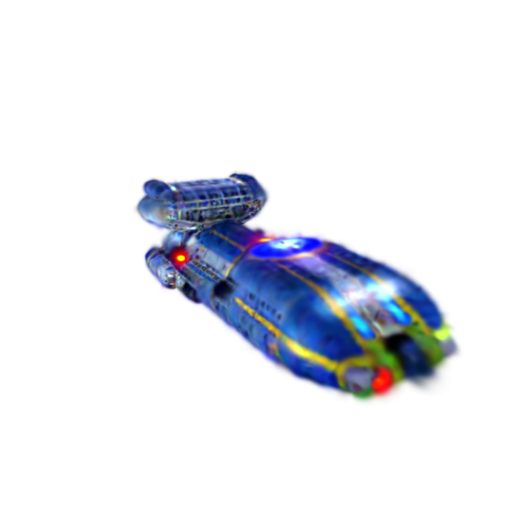}  &
                \includegraphics[width=0.24\linewidth, trim=60 60 20 50, clip]{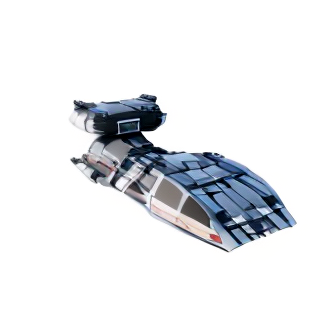} \\

                \includegraphics[width=0.24\linewidth, trim=0 0 70 70, clip]{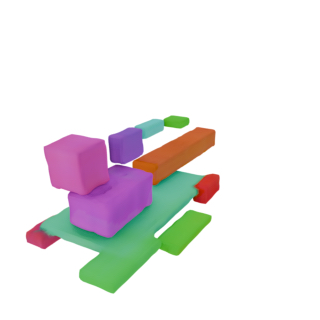} &
                \includegraphics[width=0.24\linewidth, trim=0 0 70 70, clip]{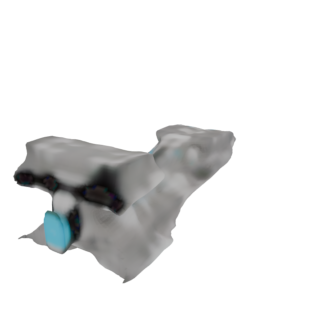} &
                \includegraphics[width=0.24\linewidth, trim=0 0 95 95, clip]{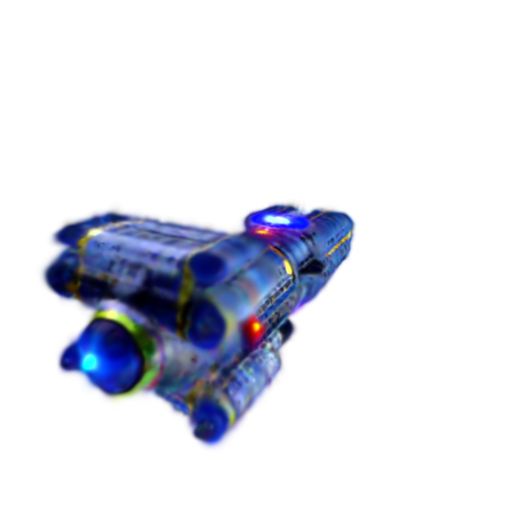}&
                \includegraphics[width=0.24\linewidth, trim=0 0 70 70, clip]{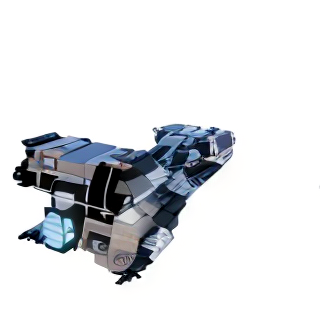}  \\

        Abstract 3D  & Spice-E & w/ SDS & w/ \ourname \\
        Guidance & & refinement & refinement
    \end{tabular}
    }
    \vspace{-8pt}
    \caption{
    \ourname{} refines outputs generated by Spice-E, in a faster and more visually appealing way than an SDS-based refinement.
    }
    \vspace{-14pt}
    \label{fig:controlled-generation}
\end{figure}

%% file: figures/edits_appearance.tex
\begin{figure}
    \centering
    \setlength{\tabcolsep}{1pt}
    {\small
    \begin{tabular}{cccc}
        \multicolumn{4}{c}{``Golden Jewelry Box'' $\longrightarrow$ ``Suede Leather Jewelry Box''} \\
        \includegraphics[width=0.22\linewidth]{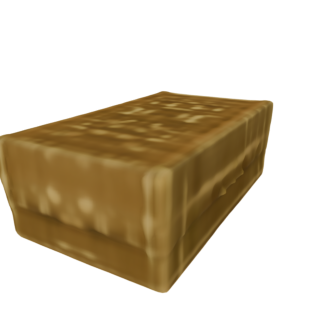} &
        \includegraphics[width=0.22\linewidth]{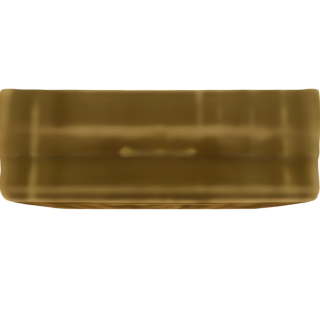} &
        \includegraphics[width=0.22\linewidth]{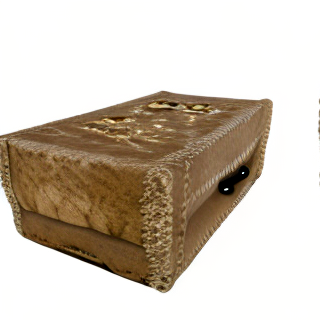} &
        \includegraphics[width=0.22\linewidth]{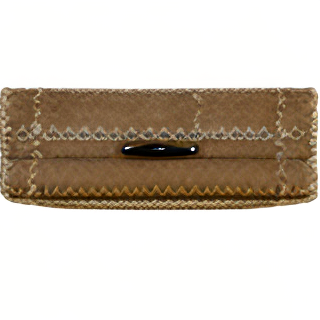} \\
        \multicolumn{4}{c}{``Glass City Tower'' $\longrightarrow$ ``Wooden City Tower''} \\
        \includegraphics[width=0.22\linewidth]{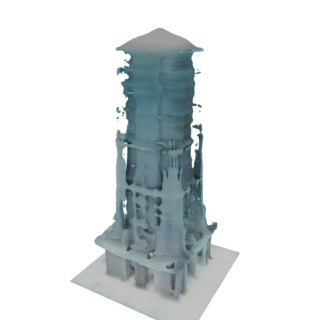} &
        \includegraphics[width=0.22\linewidth]{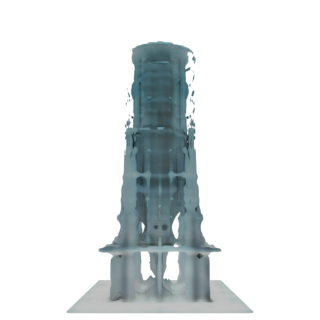} &
        \includegraphics[width=0.22\linewidth]{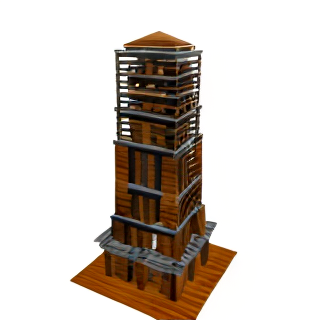} &
        \includegraphics[width=0.22\linewidth]{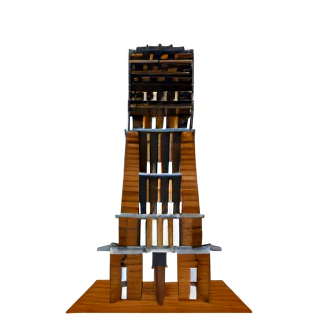} \\
        \multicolumn{4}{c}{``Leather Chair'' $\longrightarrow$ ``Leopard Print Leather Chair''} \\
        \includegraphics[width=0.22\linewidth]{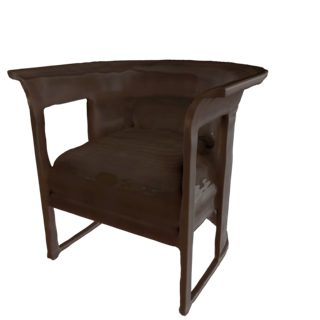} &
        \includegraphics[width=0.22\linewidth]{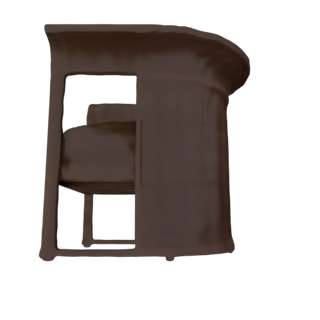} &
        \includegraphics[width=0.22\linewidth]{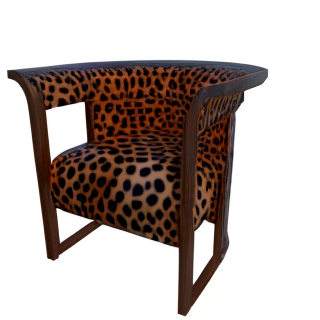} &
        \includegraphics[width=0.22\linewidth]{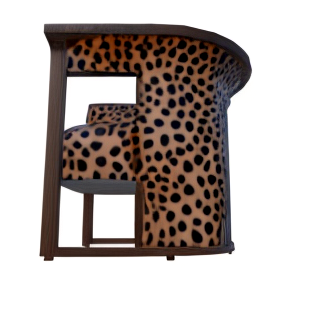} \\
        \multicolumn{2}{c}{Input} & \multicolumn{2}{c}{Appearance Edit}
    \end{tabular}
    }
    \vspace{-6pt}
    \caption{
    \ourname{} allows to edit the appearance of a Shap-E object by changing the prompt during the enhancement step. 
    }
    \vspace{-14pt}
    \label{fig:edits-appearance}
\end{figure}

%% file: 5_concolusion.tex
\vspace{-5pt}
\section{Discussion and Conclusion}
\vspace{-3pt}
We introduced \ourname, a multi-view to multi-view diffusion model that significantly enhances synthetic low-quality 3D objects, thereby improving 3D synthesis and editing pipelines. 
Our approach has two main advantages over existing multi-view generation methods. First, it relies on a coarse 3D shape, and therefore does not suffer from flat generated geometries or artifacts such as the Janus problem. Second, it can be combined with methods that operate in the latent space of the 3D generative model to provide controllability and editing capabilities.

In this work, we used Shap-E~\cite{jun2023shape} as our 3D generative model backbone, as it provides editing~\cite{chen2023shapeditor} and control~\cite{sella2024spicee} capabilities. However, our approach is applicable to other 3D generative models that encode shapes into lower-quality spaces.
One limitation of our method is its reliance on sparse-view reconstruction techniques to generate 3D objects. The reconstruction method we use does not model lighting, instead baking it into the texture. Recent advancements in sparse-view 3D reconstruction~\cite{jin2024lvsmlargeviewsynthesis, zhuang2024gtr} could potentially improve our 3D synthesis results.

We see \ourname{} as an enabler to leverage the advantages of 3D generative models, without compromising on the quality of the results. Through extensive experiments we demonstrated how \ourname{} can be integrated with various methods to achieve rich set of 3D generative capabilities. We hope that our work will pave the way for more work with native 3D generative models, leveraging their powerful generative capabilities.

%% file: 6_acks.tex
% \section*{Acknowledgement}
\section*{Acknowledgements}
We thank Matan Kleiner, Vladimir Kulikov, Ofry Livney, Rinon Gal and Yuval Alaluf for proofreading our manuscript and for their useful suggestions. 
We thank Etai Sella for his assistance in exemplifying \ourname{}'s use in refining Spice-E outputs.